\documentclass[sigconf,screen]{acmart}

\usepackage{xspace}
\usepackage{multicol}
\usepackage{multirow}
\usepackage{xcolor}
\usepackage{color, colortbl}
\usepackage[most]{tcolorbox}
\usepackage{paralist}
\usepackage{microtype}
\usepackage{balance}
\AtBeginDocument{%
  \providecommand\BibTeX{{%
    \normalfont B\kern-0.5em{\scshape i\kern-0.25em b}\kern-0.8em\TeX}}}

\setcopyright{acmlicensed}
\acmPrice{15.00}
\acmDOI{10.1145/3611643.3613879}
\acmYear{2023}
\copyrightyear{2023}
\acmSubmissionID{fse23industry-p57-p}
\acmISBN{979-8-4007-0327-0/23/12}
\acmConference[ESEC/FSE '23]{Proceedings of the 31st ACM Joint European Software Engineering Conference and Symposium on the Foundations of Software Engineering}{December 3--9, 2023}{San Francisco, CA, USA}
\acmBooktitle{Proceedings of the 31st ACM Joint European Software Engineering Conference and Symposium on the Foundations of Software Engineering (ESEC/FSE '23), December 3--9, 2023, San Francisco, CA, USA}
\received{2023-05-18}
\received[accepted]{2023-07-31}

\newcommand{\method}{\textsf{KDDT}\xspace}
\newcommand{\methodnodtm}{\textsf{\method-NoDTM}\xspace}
\newcommand{\methodwv}{\textsf{\method-Word2Vec}\xspace}
\newcommand{\company}{\textsf{Alstom}\xspace}
\newcommand{\rs}{\textsf{RWS}\xspace}
\newcommand{\packetloss}{\textsf{packet loss}\xspace}
\newcommand{\cbox}[2][!gray]{%
  \colorbox{#1}{\parbox{\dimexpr\linewidth-2\fboxsep}{#2}}%
  \vspace{-4mm}
}
\begin{document}

\newtcbtheorem{Summary}{Conclusion}{
breakable,
theorem name,
enhanced,
coltitle=black,
top=0.1in,
left=0in,
attach boxed title to top left=
{xshift=1.5em,yshift=-\tcboxedtitleheight/2},
boxed title style={size=small,colback=gray!60}
}{summary}

\title{\method: Knowledge Distillation-Empowered Digital Twin for Anomaly Detection}

\author{Qinghua Xu}
\orcid{0000-0001-8104-1645}

\affiliation{%
\institution{Simula Research Laboratory and University of Oslo}
\city{Oslo}
\country{Norway}}
\email{qinghua@simula.no}

\author{Shaukat Ali}
\orcid{0000-0002-9979-3519}
\affiliation{%
\institution{Simula Research Laboratory and Oslo Metropolitan University}
\city{Oslo}
\country{Norway}}
\email{shaukat@simula.no}

\author{Tao Yue}
\orcid{0000-0003-3262-5577}
\affiliation{%
\institution{Simula Research Laboratory}
\city{Oslo}
\country{Norway}}
\email{tao@simula.no}

\author{Zaimovic Nedim }
\orcid{0009-0005-0391-6473}
\affiliation{%
\institution{Alstom Rail Sweden AB}
\city{Västerås}
\country{Sweden}}
\email{nedim.zaimovic@alstomgroup.com}

\author{Inderjeet Singh}
\orcid{0009-0000-9022-9850}
\affiliation{%
\institution{Alstom Rail Sweden AB}
\city{Västerås}
\country{Sweden}}
\email{singh.inderjeet@alstomgroup.com}

\renewcommand{\shortauthors}{Qinghua Xu, Shaukat Ali, Tao Yue, Zaimovic Nedim, and Inderjeet Singh}

\begin{abstract}
Cyber-physical systems (CPSs), like train control and management systems (TCMS), are becoming ubiquitous in critical infrastructures. As safety-critical systems, ensuring their dependability during operation is crucial. Digital twins (DTs) have been increasingly studied for this purpose owing to their capability of runtime monitoring and warning, prediction and detection of anomalies, etc. However, constructing a DT for anomaly detection in TCMS necessitates sufficient training data and extracting both chronological and context features with high quality. Hence, in this paper, we propose a novel method named \method for TCMS anomaly detection. \method harnesses a language model (LM) and a long short-term memory (LSTM) network to extract contexts and chronological features, respectively. To enrich data volume, \method benefits from out-of-domain data with knowledge distillation (KD). We evaluated \method with two datasets from our industry partner \company and obtained the F1 scores of 0.931 and 0.915, respectively, demonstrating the effectiveness of \method. We also explored individual contributions of the DT model, LM, and KD to the overall performance of \method, via a comprehensive empirical study, and observed average F1 score improvements of 12.4\%, 3\%, and 6.05\%, respectively.

\end{abstract}

\ccsdesc[500]{Software and its engineering~Maintaining software}
\ccsdesc[500]{Security and privacy~Intrusion detection systems}
\ccsdesc[500]{Computer systems organization~Embedded systems}
\ccsdesc[500]{Computer systems organization~Sensors and actuators}
\ccsdesc[500]{Computer systems organization~Embedded and cyber-physical systems}
\ccsdesc[500]{Computing methodologies~Neural networks}

\keywords{digital twin, knowledge distillation, anomaly detection, Train Control and Management System}


\maketitle

\section{Introduction}
\label{sec:intro}

Cyber-physical Systems (CPSs) play a vital role in Industry 4.0. Various CPSs have been deployed in critical infrastructures, e.g., water treatment plant~\cite{xu_digital_2021,xu_digital_2023} and elevator systems ~\cite{xu_uncertainty-aware_2022, liping,liping-tosem}, whose safe operation concerns our daily lives. One such safety-critical CPS is railway systems (\rs). Our industrial partner \company manufactures intricate \rs, which involves not only vehicles and tracks (the physical part) but also network communication inside the vehicles themselves and across them (the cyber part). \rs get increasingly complex when they become more heterogeneous and integrated, i.e., with more software systems added to provide rich functionalities. However, such complexity renders \rs susceptible to broader threats (e.g., anomalies~\cite{anushiya_comparative_2021} and software faults~\cite{zafar_model-based_2021}). In particular, anomalies are one of the most severe threats that might lead to system failures or even catastrophic consequences~\cite{islam_novel_2022}.

Inspired by their success in computer vision and natural language processing (NLP)~\cite{DONG2021100379}, machine learning methods have been widely applied to address various anomaly detection tasks in \rs ~\cite{islam_novel_2022,h_zhao_fault_2017,guzman_data-driven_2018}. However, training these methods involves collecting abundant data from \rs, which might interfere with the safe operation of \rs. To reduce such interference, digital twin (DT), as a novel technology, has been intensively studied for CPSs anomaly detection~\cite{xu_digital_2021,xu_digital_2023,xu_uncertainty-aware_2022,jones2020characterising}. Specifically, a DT can be considered as a digital replica of a CPS and enables rich functionalities with this replica instead of the real CPS. Early DTs are predominantly based on software/system models, requiring manual effort from domain experts~\cite{eckhart_digital_2019}. To mitigate this challenge, machine learning methods are increasingly used to enable data-driven DT constructions, which require much less manual effort and domain knowledge while performing accurate anomaly detection in CPSs ~\cite{xu_digital_2021,xu_digital_2023,xu_uncertainty-aware_2022}. 

However, to the best of our knowledge, no previous works have focused on building data-driven DTs for network anomaly detection in \rs. In this paper, we focus on detecting \packetloss anomalies on IP networks in the Train Control and Management System (TCMS). When a \packetloss anomaly occurs, a certain pattern is exhibited where certain signal values abruptly drop to 0 and rebound quickly. Such network anomalies can induce misjudgment and incorrect reactions in the control unit of \rs, leading to unexpected and risky behaviours of \rs. Despite its significance, constructing a DT for network anomaly detection remains unsolved due to the following two challenges listed below.

\noindent\textbf{Challenge 1: High data complexity.} Datasets for network anomaly detection are packets arranged in chronological order. The packets' contents and order reflect the network state and the oblivion of either aspect diminishes the performance of the DT. Sequential models such as recurrent neural networks can extract chronological features, while the packets' content features are intrinsically more difficult to extract due to their textual format. Unlike structural data, textual data does not follow a fixed format or schema and can vary greatly in length. To extract high-quality features from textual data, both the syntax and semantics of the data should be understood by the feature extractor without ambiguity. 
\\ 
\noindent\textbf{Challenge 2: Training data insufficiency.} Data-driven DT construction entails training with abundant in-domain labelled data, i.e., data related to the network \packetloss. However, such datasets are difficult to collect since anomalies during the operation of safety-critical systems like \rs rarely occur. Moreover, detecting anomalies manually is non-trivial. The presence of packet dropping can be difficult to detect without rich domain knowledge since normal network fluctuations might resemble an anomaly to a great extent. Such resemblance prevents non-experts, without sufficient training, from manually labelling the packets.

To address the aforementioned challenges, we propose a novel DT-based method \method for network anomaly detection in TCMS. To tackle the first challenge, \method extracts features from packets' contents and order  with a contextualized language model (LM) and an LSTM, respectively~\cite{mimura2018reading,yu2019review}.  The second challenge is the insufficiency of training data. We circumvent this challenge by distilling knowledge from the out-of-domain (OOD) dataset as a supplement, which is relatively cheaper than the in-domain (ID) dataset acquisition. Specifically, we pretrain a Variational Autoencoder (VAE) to encode and reconstruct a network packet from OOD data. We posit a well-trained VAE possesses rich knowledge of encoding network packets with high quality, which can be leveraged by KD to supplement the ID dataset. Similar to the popular pretraining+fine-tuning paradigm, KD is also a transfer learning technique. However, KD asks the pretrained VAE to act like a teacher rather than merely providing a set of better-initialized parameters for \method. The key advantage of KD is a reduction in model complexity, which requires fewer resources and can potentially mitigate the risk of overfitting.

We evaluated \method with two TCMS network packet datasets from \company. \method achieves average F1 scores: 0.931 and 0.915. Moreover, we posit that the benefits of \method's sub-components extend beyond the scope of network anomaly detection. To investigate their individual contributions, we studied the effectiveness of DTM, LM, and KD. Evaluation results show that the absence of DTM, LM or KD leads to decreases of 12.4\%, 3\%, and 6.05\%, respectively, in terms of the average F1 score on both datasets.

\section{Industrial Context}

\label{sec:indus}
\begin{figure}
    \centering
    \includegraphics[width=\linewidth]{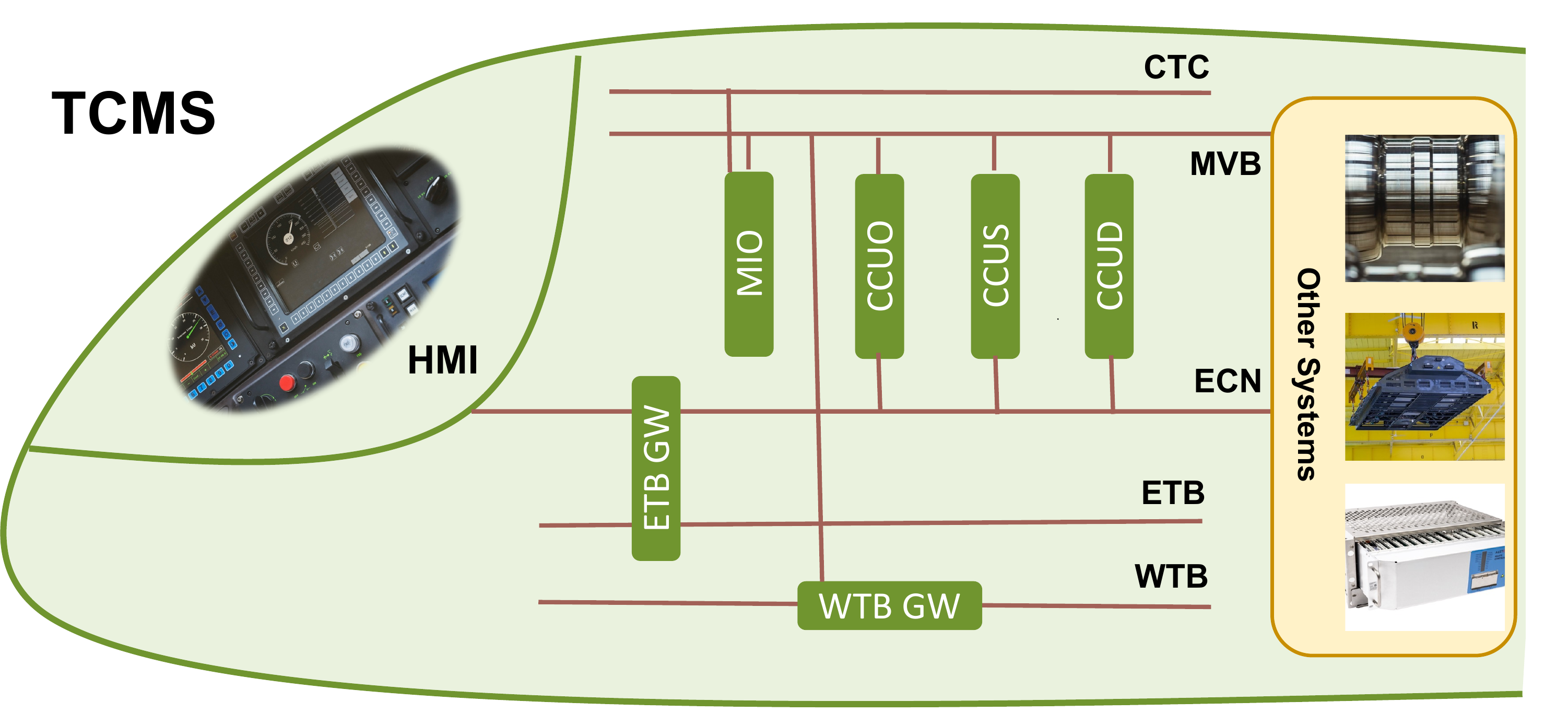}
    \caption{Overview of TCMS. MVB, ECN, ETB, and WTB stand for Multi-function Vehicle Bus, Ethernet Consist Network, Ethernet Train Bus, and Wired Train Bus, respectively. WTB GW and ETB GW are two gateways. CTC, MIO, and HMI represent Conventional Train Control, Modular Input/Output, Human Machine Interface, respectively. CCUs are control unit, including CCUO, CCUS, and CCUD. }
    \label{fig:tcms}
\end{figure}

\company Rail Sweden AB is the second largest company in the rail industry, with over 150 000 vehicles in service. One essential train component is TCMS, a high-capacity infrastructure that controls communications among different subsystems on the train and between the train and the ground\cite{zafar_model-based_2021}. Figure \ref{fig:tcms} depicts the overview of the TCMS. Four buses and two gateways constitute the backbone of TCMS, namely Multi-function Vehicle Bus (MVB), Ethernet Consist Network (ECN), Ethernet Train Bus (ETB), Wired Train Bus (WTB), WTB GW, and ETB GW. The MVB and ECN buses connect various internal devices and other subsystems, while the WTB and ETB buses connect multiple vehicles. Conventional Train Control (CTC) implements hard-wire logic in the control system. The Modular Input/Output (MIO) devices tackle Input/Output analog signals. Human Machine Interfaces (HMIs) provide control interfaces for train drivers. TCMS relies on Central Control Units (CCUs) to implement various train functions. CCUO, CCUS, and CCUD control the basic, safety-critical, and diagnostic functions, respectively. Each function is performed with one or multiple devices and buses in TCMS.

Take the standstill determination process as an example. Multiple train functions rely on accurately assessing standstill, such as door control. Concretely, the TCMS communicates with two sub-systems: the Brake Control Units (BCUs) and the Traction Control Units (TCUs). The BCUs measure the axles' speed and inform the CTC about the standstill condition by setting specific digital inputs on the MIO. In addition to the BCU's hardware-based standstill determination, the TCMS uses the speed of each driven axle, which is sent by the TCUs via the Multifunction Vehicle Bus (MVB), for the software-based standstill determination. This signal is then transmitted to the Door Control Units (DCUs) and other functions via MVB/IP. Such processes entail packet exchanges in the TCMS network, which is vulnerable to \packetloss anomalies. This paper aims to develop an effective DT for anomaly detection, facilitating the safety check of \rs and reducing future \packetloss risks.

\section{Methodology}
\label{sec:metho}
\begin{figure}[h]
  \centering
  \includegraphics[width=\linewidth]{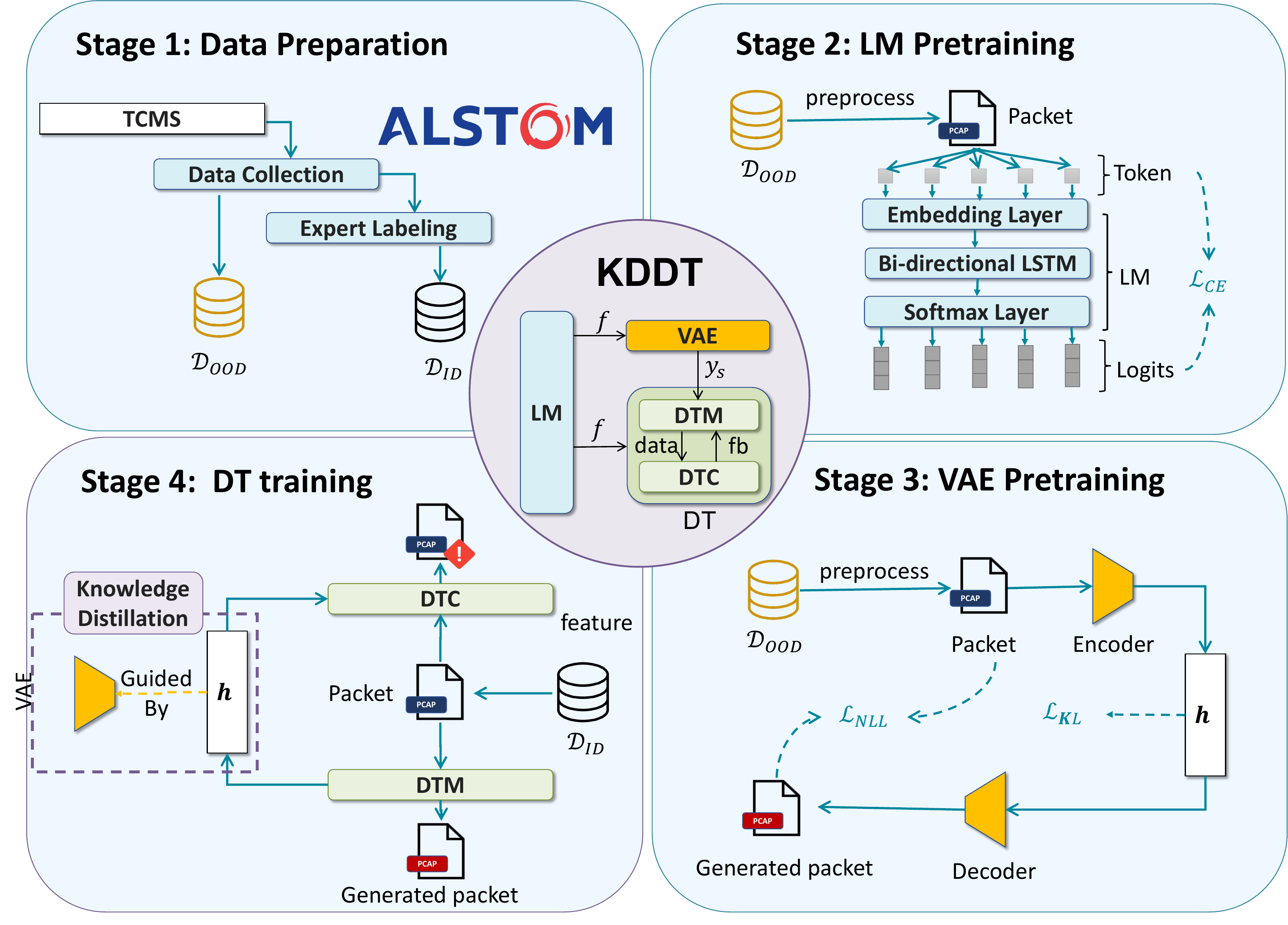}
  \caption{\method's model structure and training workflow. $\mathcal{D}_{OOD}$ and $\mathcal{D}_{ID}$ are out-of-domain and in-domain datasets.}
  \label{fig:overview}
\end{figure}
As depicted in the center circle of Figure \ref{fig:overview}, \method has three major components: a pretrained LM, a pretrained VAE, and a DT. \textit{A pretrained LM} extracts contextualized features $f$ from textual data (i.e., network packet), which are leveraged by \textit{DT} for anomaly detection. \textit{DT} comprises a digital twin model (DTM) and a digital twin capability (DTC). DTM simulates the TCMS, while DTC implements one or more functionalities of DT, e.g., anomaly detection in our context. The interaction between DTM and DTC is bidirectional, where DTM sends information about the system state to DTC and receives feedback from DTC, e.g., anomaly alerts. To further improve the DT performance, we utilize the pretrained VAE as a teacher model to guide DTM's training.

The four boxes of Figure \ref{fig:overview} illustrate the training workflow of \method: data preparation, LM pretraining, VAE pretraining, and DT training. In \textit{Stage 1}, we first prepare both out-of-domain $\mathcal{D}_{OOD}$ and in-domain $\mathcal{D}_{ID}$ datasets. As mentioned in Section \ref{sec:indus}, we collect anomaly-unrelated data as $\mathcal{D}_{OOD}$, which can be leveraged to pretrain some universal models such as LM (Stage 2) and VAE (Stage 3). In-domain dataset $\mathcal{D}_{ID}$, on the other hand, are directly related to anomalies, which can be used to train DT for anomaly detection (Stage 4). In \textit{Stage 2}, we pretrain a contextualized LM with the $\mathcal{D}_{OOD}$ dataset to extract inner-packet context features by predicting the next token. For instance, given the context of "\rs is an important Cyber-physical", a well-trained LM can extract a discrete vector to represent the context and use the vector to predict the next token, i.e., "system" with a high probability. In \textit{Stage 3}, we design a VAE comprising an encoder and a decoder and pretrain it with packets from the $\mathcal{D}_{OOD}$ dataset. The encoder extracts context features with the pretrained LM and encodes them into a hidden vector (\textit{h}). The decoder then takes \textit{h} as the input and attempts to reconstruct the original network packets. A well-trained VAE can effectively encode a network packet into a high-quality hidden vector and reconstruct the packet with high fidelity. In \textit{Stage 4}, we train DTM and DTC of \method with the $\mathcal{D}_{ID}$ dataset but under the guidance of the pretrained VAE. In particular, DTM and the pretrained VAE both aim to encode a given packet with high quality. VAE contains richer knowledge due to its complex architecture and pretraining on the $\mathcal{D}_{OOD}$ dataset. Consequently, the hidden vectors produced by the pretrained VAE convey richer information, which can be used as a soft target (as opposed to a hard target, i.e., ground truth target) to guide the DTM training process.

\subsection{Data Preparation}
\label{subsec:data}
As depicted in Figure \ref{fig:overview}, Stage 1, we sequentially perform two processes in the data preparation stage: data collection and domain expert labelling. 

\noindent\textbf{Data collection.} During operation, TCMS communicates with various devices by exchanging large numbers of packets per second. Engineers in \company deployed Wireshark~\cite{lamping2004wireshark} to monitor such exchanges and capture packets as \textit{.PCAP} files. Let $X_i$ be the packet captured at timestep $i$. We collect out-of-domain dataset $\mathcal{D}_{OOD}=[X_0,X_1,...,X_{N_{OOD}-1} ]$ and in-domain dataset $\mathcal{D}_{ID}=[X_0,X_1,...,X_{N_{ID}-1} ]$, where $ {N_{OOD}}$ and ${N_{ID}}$ represent the dataset sizes. The $\mathcal{D}_{ID}$ dataset refers to any data related to the anomaly detection task, i.e., suspicious anomalous data. $\mathcal{D}_{OOD}$ dataset is task-agnostic, which includes any network data collected in TCMS.

\noindent\textbf{Domain expert labelling.} \method, as a supervised learning method, requires labelled data for accurate network anomaly detection. In our context, we study the phenomenon of \packetloss. Hence we assign an abnormal label to a packet if it is experiencing a \packetloss incident. We ask \company to provide us with labelled packet data. They manually examined the signal logs with the help of an in-house built monitoring tool named DCUTerm. DCUTerm displays signal changes over time, providing a rough time boundary for \packetloss anomalies, which was further analyzed and narrowed down by checking the Wireshark packets directly. We formally denote the label for $X_i$ as  $y_i \in \{0,1\}$, which takes the value of $1$ when a \packetloss incident occurs.

\subsection{LM Pretraining}
\label{subsec:embed}
The quality of feature engineering dramatically influences the performance of machine learning models. One of the most salient features of the packets is the inner-packet context features, which represent the semantics and syntax of the packet content. 
We aim to pretrain a contextualized LM to extract such features from packet contents. The LM treats the raw packet content as natural language text and extracts semantic and syntactic features automatically. We will demonstrate the model structure of LM in Section \ref{subsubsec:lm_model} and the loss function in Section \ref{subsubsec:lm_loss}.

\subsubsection{LM model structure}
\label{subsubsec:lm_model}
As shown in Figure \ref{fig:overview}, the LM consists of three layers:
~\\
\noindent\textbf{Embedding layer.} We first build a vocabulary $\mathcal{V}$ for all the packet tokens and randomly initialize an N-dimensional embedding vector for each token. Given a training sample $(ctx,tgt)$, the embedding layer takes the context sequence $ctx$ as input and fetches the corresponding embedding for each token inside this sequence, as shown in Equation \ref{eq:embed}. $L$ denotes the length of the context.
    \begin{equation}
        \label{eq:embed}
        X=Embed(ctx)=[x_0,x_1,...,x_{L-1}]
    \end{equation}

\noindent\textbf{Bi-directional LSTM layer.} We feed the embedded context $X$ into a bi-directional LSTM ~\cite{yu2019review} for context information extraction:
    \begin{equation}
        \label{eq:bi-lstm}
        h=LSTM(\overset{\leftarrow}{X}) + LSTM(\overset{\rightarrow}{X})
    \end{equation}

\noindent\textbf{Softmax layer.} This layer transforms the output of the bi-directional LSTM into logits $z$ that sum up to be 1 as in Equation \ref{eq:soft}:
    \begin{equation}
        \label{eq:soft}
        \mathrm{z} = \mathrm{softmax}(h) = \frac{\exp(h_i)}{\sum_{j=1}^N \exp(h_j)}
    \end{equation}

\subsubsection{Loss Calculation}
\label{subsubsec:lm_loss}
 The output of LM is a probability distribution vector, whose element represents each token's probability of appearing at the next position. In Equation \ref{eq:ce}, We calculate Cross Entropy Loss $\mathcal{L}_{CE}$ by comparing logits $\mathrm{z}$ (Equation \ref{eq:soft}) with true values $tgt$, which is minimized to train the LM. $|\mathcal{V}|$ denotes the vocabulary size. 
    \begin{equation}
        \label{eq:ce}
        \mathcal{L}_{CE} = -\sum_{i=1}^{|\mathcal{V}|} \mathrm{tgt}_i \log \mathrm{z}_i
    \end{equation}

\subsection{VAE Pretraining}

To leverage the $\mathcal{D}_{OOD}$ dataset, we train VAE as illustrated in Stage 3, Figure \ref{fig:overview}. The underlining hypothesis of VAE is that each packet is an observed data point generated based on a hidden vector that summarises this packet. VAE aims to find this hidden vector with an encoder and evaluate its quality by reconstructing the packet with the decoder. Higher reconstruction quality indicates better encoding capability. Given a packet $X_i\in \mathcal{D}_{OOD}$, the encoder extracts a hidden state vector, with which the decoder reconstructs a packet similar  to (but necessarily the same as) $X_i$. We will present the details about the encoder and decoder in Section \ref{subsubsec:encoder} and Section \ref{subsubsec:decoder}, respectively.

\label{subsec:dtm_pretraining}
\begin{figure*}
    \centering
    \includegraphics[width=0.9\linewidth]{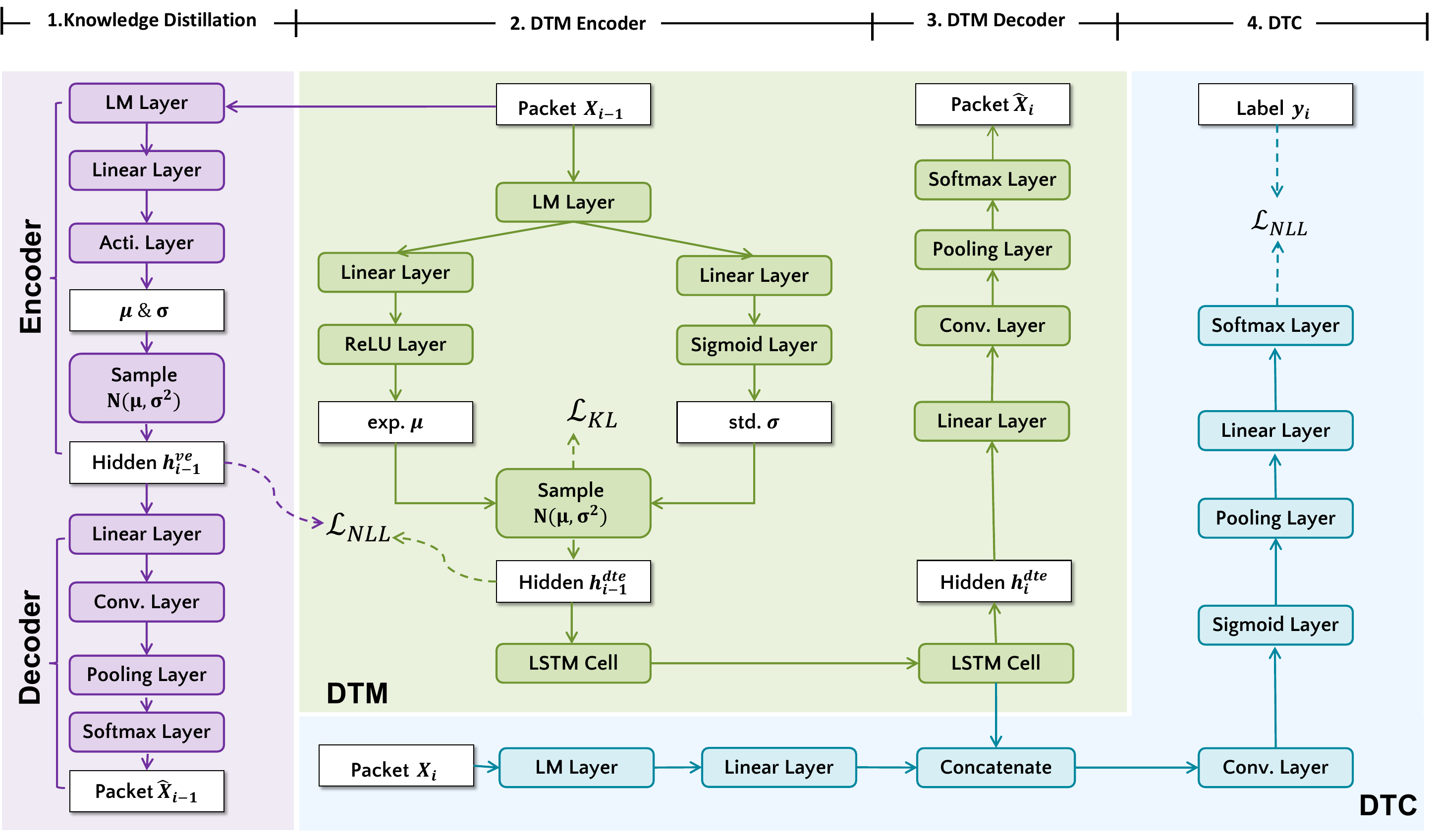}
    \caption{Model Structure of \method}
    \label{fig:dt}
\end{figure*}
\subsubsection{VAE Encoder}
\label{subsubsec:encoder}
The encoder aims to induce the hidden vector from $\mathcal{D}_{OOD}$ packets, with an assumption that the hidden vector follows a Gaussian distribution $N(\mu,\sigma^2)$. The high expressiveness of the Gaussian distribution allows it to describe many phenomena in the real world. According to~\cite{pinheiro_cinelli_variational_2021}, the Gaussian distribution assumption allows VAE to utilize the reparameterization trick, which enhances training efficiency without reducing its fitting capability. VAE first approximates the Gaussian distribution $N(\mu,\sigma^2)$ by computing $\mu$ and $\sigma$ with two neural network models. Then we sample a hidden vector from $N(\mu,\sigma^2)$ with the reparameterization trick. The upper part of the purple box in Figure \ref{fig:dt} illustrates the internal layers of the VAE encoder:
~\\
\noindent\textbf{LM layer.} We utilize the trained LM from Stage 2 to convert the packet input $X$ into embedding vectors as in Equation \ref{eq:dtm_lm}.
    \begin{equation}
        \label{eq:dtm_lm}
        X=LM(X)
    \end{equation}
\noindent\textbf{Linear layers for $\mu$ and $\sigma$.} We perform two separate linear transformations for $\mu$ and $\sigma$ as in Equations \ref{eq:mu_lin} and \ref{eq:sig_lin} respectively, where $W$ and $b$ are weight and bias matrices. 
    \begin{equation}
        \label{eq:mu_lin}
        \mu=W_{\mu}X+b_{\mu}
    \end{equation}
    \begin{equation}
        \label{eq:sig_lin}
        \sigma=W_{\sigma}X+b_{\sigma}
    \end{equation}
\noindent\textbf{Activation layers for $\mu$ and $\sigma$.} We introduce non-linearity into the model by adding an activation layer. Following the common practice in~\cite{pinheiro_cinelli_variational_2021}, we choose ReLU and sigmoid as the activation function for $\mu$ and $\sigma$, respectively (Equations \ref{eq:acti_mu} and \ref{eq:acti_sig}). 
    \begin{align}
        \label{eq:acti_mu}
        \mu&=ReLU(\mu)\\
        \label{eq:acti_sig}
        \sigma &= Sigmoid(\sigma)
    \end{align}
\noindent\textbf{Sampling layer.} Given the expectation $\mu$ and standard deviation $\sigma$ for a Gaussian distribution, it is trivial to sample a hidden vector $h_{vE}$ from it. However, the practice of sampling breaks the gradient propagation chain, which is the foundation of deep learning model training. Therefore, VAE uses a reparametrization trick to address this problem as in Equation \ref{eq:rep}, where $\epsilon$ is random noise. 
    \begin{equation}
        \label{eq:rep}
        h_{vE}=\mu+\epsilon*\sigma
    \end{equation}

\subsubsection{VAE Decoder}
\label{subsubsec:decoder}
The decoder aims to reconstruct the packet from the hidden vector $h_{vE}$. We build it as a convolution neural network consisting of the following layers:
 ~\\

\noindent\textbf{Linear layer.} Equation \ref{eq:decoder_l} shows that we first perform a linear transformation on the hidden vector $h_{vE}$, where $W_d$ and $b_d$ are weight and bias matrices.
    \begin{equation}
        \label{eq:decoder_l}
        h_{vD}=W_d h_{vE}+b_d
    \end{equation}
\noindent\textbf{Convolution layer.} After the linear transformation, we use a convolution layer to capture spatial features in $h_{vd}$:
    \begin{equation}
        \label{eq:conv}
        h_{vD}=Conv(h_{vD})
    \end{equation}
\noindent\textbf{Pooling layer.} We add a maximum pooling layer to increase the fitting capability of the decoder as in Equation \ref{eq:pool}.
    \begin{equation}
        \label{eq:pool}
        h_{vD}=max\_pool(h_{vD})
    \end{equation}
\noindent\textbf{Softmax layer.} Finally, we project the reconstructed vectors into the packet space with a softmax layer as in Equation \ref{eq:decoder_sof}.
    \begin{equation}
        \label{eq:decoder_sof}
        \hat{X}=softmax(h_{vD})
    \end{equation}

\subsubsection{Loss Calculation}
\label{subsubsec:loss}
As pointed out in the literature~\cite{pinheiro_cinelli_variational_2021}, an overall loss should be minimized to train VAE, consisting of a KL divergence loss for the encoder and a maximum likelihood loss (MLL) for the decoder (Equation \ref{eq:loss}). 
\begin{equation}
    \label{eq:loss}
    \mathcal{L}_{VAE}=\mathcal{L}_{KL}+\mathcal{L}_{MLL}
\end{equation}
We compute the KL divergence loss by comparing the hidden vector distribution with a standard Gaussian distribution:
\begin{align}
    \mathcal{L}_{KL}&=KL(N(\mu,\sigma^2)||N(0,1)) \notag \\
    &=-\frac{1}{2} \sum_{i=1}^k \left(1 + \log \sigma_i^2 - \mu_i^2 - \sigma_i^2 \right)
    \label{eq:kl}
\end{align}
The maximum likelihood loss aims to assess how close the reconstructed packet $\hat{X}_{i}$ is to the real $\hat{X}_{i}$. As shown in Equation \ref{eq:mll}, we calculate the cross entropy as the MLL loss.
\begin{equation}
\label{eq:mll}
\mathcal{L}_{MLL} = -\sum_{j=1}^{n}X_{i+1}^j\log(\hat{X}_{i+1}^j)
\end{equation}

\subsection{DT Training}
The last stage in Figure \ref{fig:overview} is DT training, which takes advantage of the collected data, pretrained LM and VAE from Stages 1, 2 and 3, respectively. Figure \ref{fig:dt} presents model details of DT, comprising a DTM, a DTC, and a KD module.
\subsubsection{DTM Structure}
\label{subsubsec:dtm}
DTM simulates the TCMS network by predicting the subsequent packet. As mentioned in Section \ref{sec:intro}, the predictive performance of anomaly detection hinges on extracting inner-packet context and inter-packet chronological features. A well-trained VAE can effectively extract the inner-packet context feature, but the inter-packet chronological feature is not considered in the VAE structure. Therefore, we adapt the VAE from Stage 3 by combining VAE and LSTM to extract both features.

At timestep $i-1$, DTM sequentially invokes three sub-models to predict a packet $X_i$: an encoder, an LSTM and a decoder (Figure \ref{fig:dt}).  
~\\
\noindent\textbf{DTM Encoder.} The internal structure of the encoder is identical to the VAE encoder. At timestep $i-1$, the encoder encodes the input packet from previous timestep $X_{i-1}$ into a hidden state vector $h_{i-1}^{dtmE}$ (Equation \ref{eq:dtm_encoder}). See details of the encoder in Section \ref{subsec:dtm_pretraining}. 
\begin{equation}
    \label{eq:dtm_encoder}
    h_{i-1}^{dtmE}=encoder(X_i)
\end{equation}
However, we reduce the complexity of the DTM encoder by using a smaller hidden vector size compared to the VAE encoder. KD can benefit from a complexity discrepancy between the teacher and student models~\cite{clark2018semi}.

%
\noindent\textbf{DTM LSTM.} We utilize an LSTM to capture the inter-packet chronological features. The LSTM takes the hidden vector $h_{i-1}^{dtmE}$ as input and outputs a hidden vector $h_i^{dtmE}$ for timestep $i$ as in Equation \ref{eq:dtm_lstm}. 
\begin{equation}
    \label{eq:dtm_lstm}
    h_i^{dtmE}=LSTM(h_{i-1}^{dtmE})
\end{equation}
\noindent\textbf{DTM Decoder.} The decoder aims to generate a packet $\hat{X}_i$ with the hidden vector $h_i$ as in Equation \ref{eq:dtm_decoder}. 
\begin{equation}
    \label{eq:dtm_decoder}
    \hat{P}_i=dtm\_decoder(h_i^{dtmE})
\end{equation}
\subsubsection{DTC Structure}
\label{subsubsec:dtc}
DTC performs anomaly detection with the input from the packet $X_i$ and the hidden vector $h_i^{dtmE}$ from DTM. $h_i^{dtmE}$ is an indicator of the system state based on historical data, whereas $X_i$ contains information about the signal values in the TCMS. As shown in Figure \ref{fig:dt}, DTC consists of the following layers:
 ~\\
\noindent\textbf{LM layer.} Similar to the LM layer in DTM, we use the pretrained LM to convert packet $P_i$ into embedding vectors as in Equation \ref{eq:dt_lm}.
    \begin{equation}
        \label{eq:dt_lm}
        X_i=LM(X_i)
    \end{equation}
\noindent\textbf{Linear layer.} This layer linearly transforms the embeddings into the hidden space of DTM as in Equation \ref{eq:dt_l}, where $W_{l1}$ and $b_{l1}$ are weight matrices.
    \begin{equation}
        \label{eq:dt_l}
        h_i^{dtc}=W_{l1} X_i +b_{l1}
    \end{equation}
\noindent\textbf{Convolution layer.} We then concatenate hidden vectors from DTM $h_{i}^M$ and DTC $h_{i}^{dtc}$ and feed them into a convolution layer as in Equation \ref{eq:dt_c}.
    \begin{equation}
        \label{eq:dt_c}
        h_i^{dtc}=Conv([h_{i}^{dtmE},h_{i}^{dtc}])
    \end{equation}
\noindent\textbf{Sigmoid layer.} Non-linearity is introduced with a sigmoid activation function as in Equation \ref{eq:dt_s}.
    \begin{equation}
        \label{eq:dt_s}
        h_i^{dtc}=Sigmoid(h_i^{dtc})
    \end{equation}
\noindent\textbf{Pooling layer.} We reduce the dimensionality of $h_i^{dtc}$ with a pooling layer as in Equation \ref{eq:dt_p}.
    \begin{equation}
        \label{eq:dt_p}
        h_i^{dtc}=max\_pool(h_i^{dtc})
    \end{equation}
\noindent\textbf{Linear layer.} We transform the hidden vector $h_i^{dtc}$ into the output space as in Equation \ref{eq:dt_ll}, where $W_{l2}$ and $W_{l2}$ are weight matrices.
    \begin{equation}
        \label{eq:dt_ll}
        o_i^{dtc}=W_{l2} h_i^{dtc} +b_{l2}
    \end{equation}
\noindent\textbf{Softmax layer.} Finally, we perform a softmax operation to convert the output $o_i$ into logits as in Equation \ref{eq:dt_sl}.
    \begin{equation}
        \label{eq:dt_sl}
        z_i^{dtc}=softmax(o_i^{dtc})
    \end{equation}

\subsubsection{Knowledge Distillation}
\label{subsubsec:kd}
The pretrained VAE learns from extra OOD data about extracting inner-packet context features. To make use of the pretrained models, many researchers adopt a rather intuitive strategy to use the pretrained parameters directly and fine-tune them with ID datasets. KD, however, tailors the complex teacher model into a less complex student model to focus on a specific context, i.e., anomaly detection in our context. Instead of direct parameter sharing, the student model uses the hidden vectors produced by the teacher model as a soft target and optimizes its own parameters to encode similar vectors as the teacher model. Figure \ref{fig:dt} shows the pretrained VAE takes the packet from the previous timestep $X_{i-1}$ as input and produces a hidden vector $h_{i-1}^{vE}$. The DTM encoder uses $h_{i-1}^{vE}$ as a soft target and calculates a cosine similarity loss:
\begin{equation}
    \label{eq:kd}
    \mathcal{L}_{KD}=\frac{h_{i-1}^{vE} \cdot h_{i-1}^{dtmE}}{|h_{i-1}^{vE}| \times |h_{i-1}^{dtmE}|}
\end{equation}

\subsubsection{Loss Calculation}
\label{subsubsec:dt_loss}
The overall loss to train \method comprises the ground truth loss and KD loss as depicted in Equation \ref{eq:dt_loss}.
\begin{equation}
    \label{eq:dt_loss}
    \mathcal{L}=\mathcal{L}_{GT}+\mathcal{L}_{KD}
\end{equation}
$\mathcal{L}_{KD}$ is calculated as in Equation \ref{eq:kd}, while $\mathcal{L}_{GT}$ entails loss calculation on both DTM and DTC (Equation \ref{eq:loss_gt}).
\begin{equation}
    \label{eq:loss_gt}
    \mathcal{L}_{GT}=\mathcal{L}_{DTM}+\mathcal{L}_{DTM}
\end{equation}

Since DTM resembles VAE in the model structure, we calculate the DTM loss $\mathcal{L}_{DTM}$ similarly as a sum of the KL loss and MLL loss (Equation \ref{eq:dtm_loss}). Detailed calculation is in Equations \ref{eq:kl} and \ref{eq:mll}.
\begin{equation}
    \label{eq:dtm_loss}
    \mathcal{L}_{DTM}=\mathcal{L}_{KL}+\mathcal{L}_{MLL}
\end{equation}

We compute a commonly-used cross entropy loss for DTC as: 
\begin{equation}
    \mathcal{L}_{CE} = -\sum_{i=1}^{N} \mathrm{y}_i \log \mathrm{z}_i^{dtc}
\end{equation}

\section{Experiment Design}
\label{sec:experi_desig}
To evaluate \method, we propose four research questions (RQs) in Section \ref{subsec:rq}. Section \ref{subsec:industry} presents details on the subject system and collected dataset. In Section \ref{subsec:em} and Section \ref{subsec:es}, we discuss the evaluation metrics and experiment setups.

\subsection{Research Questions}
\label{subsec:rq}
\textbf{RQ1:} Is \method effective in detecting anomalies in TCMS?
\textbf{RQ2:} Is DTM effective in improving DTC performance?
\textbf{RQ3:} Is LM effective in extracting inner-packet features?
\textbf{RQ4:} Is KD effective in improving DTM's encoding capability?
    
In RQ1, we investigate the effectiveness of \method in anomaly detection. With RQ2 -- RQ4, we delve into the contribution of each sub-component (i.e., DTM, LM and KD) of \method to its overall effectiveness. Specifically, in RQ2, we compare the effectiveness of \method with/without DTM. In RQ3, we study the effectiveness of LM and compare it to Word2vec, another text feature extraction method effective for packet-related tasks~\cite{goodman_packet2vec_2020}. In RQ4, we compare the effectiveness of \method with/without KD. 

\subsection{Industrial Subject System}
\label{subsec:industry}
Our industrial partner \company provided us with packet data captured in their TCMS. Anomalies, i.e., \packetloss incidents, occur on different devices at different times due to various environmental uncertainties. They collected packet data and manually identified two periods of time when anomalies tend to appear more frequently, i.e., 2021/02/10 09:00 pm - 09:20 pm and 2021/02/11 05:35 am - 05:42 am. As mentioned in Section \ref{subsec:data}, we collect both OOD dataset $\mathcal{D}_{OOD}$ and ID dataset $\mathcal{D}_{ID}$. We further divide $\mathcal{D}_{ID}$ into a training dataset and a testing dataset for the evaluation purpose with a ratio of 0.8:0.2. Table \ref{tab:dataset} reports key statistics about the dataset.  
\setlength{\intextsep}{10pt}
\setlength{\textfloatsep}{0pt}
\setlength{\abovecaptionskip}{0pt}
\setlength{\belowcaptionskip}{0pt}
\begin{table}[hbt]
    \centering
    \begin{tabular}{ccccc}
    \toprule
        Metric &  Dataset & $\mathcal{D}_{OOD}$ & $\mathcal{D}_{ID}$-Train & $\mathcal{D}_{ID}$-Test\\
        \midrule
        \rowcolor{gray!20}
        \cellcolor{white}\multirow{2}{*}{$\mathcal{N}$} & Day 10 & 625525  & 750631 & 187658 \\
        & Day 11 & 313007  &375609 &93903 \\\hline
        \rowcolor{gray!20}
        \cellcolor{white}\multirow{2}{*}{$\mathcal{N}_{NP}$} & Day 10 & 625525  & 713990 & 137808 \\
        & Day 11 & 313007  &339395 &70225 \\\hline
        
        \rowcolor{gray!20}
        \cellcolor{white}\multirow{2}{*}{$\mathcal{N}_{AI}$} & Day 10 & -  & 186 & 122 \\
        & Day 11 & -  &64 &21 \\\hline
        \rowcolor{gray!20}
        \cellcolor{white}\multirow{2}{*}{$\mathcal{N}_{AP}$} & Day 10 & -  &36641 & 49850 \\
        & Day 11 & - &36214 & 23678 \\\hline
        \rowcolor{gray!20}
        \cellcolor{white}\multirow{2}{*}{$\mathcal{L}_{AI}$} & Day 10 &-  & 197.00 & 407.23 \\
        & Day 11 & -  & 565.84 & 1127.52 \\\hline
        \rowcolor{gray!20}
        \cellcolor{white}\multirow{2}{*}{$\mathcal{T}_{AI}$} & Day 10 & -  &198353.88 $\mu s$ &420050.60 $\mu s$ \\
        & Day 11 & -  & 616352.65 $\mu s$ & 1207371.71 $\mu s$ \\\hline
        \bottomrule
    \end{tabular}
    \caption{Statistics about the dataset. The $\mathcal{D}_{OOD}$ dataset contains only anomaly-unrelated data, while $\mathcal{D}_{ID}$ contains anomaly-related data. $N$ represents the total number of packets in the dataset. $\mathcal{N}_{NP}$ and $\mathcal{N}_{AP}$ are the numbers of normal and abnormal packets in the dataset. $\mathcal{N}_{AI}$ is the number of anomaly incidents in the dataset. $\mathcal{L}_{AI}$ and $\mathcal{T}_{AI}$ denote the average length (number of packets in an anomaly) and time duration of anomalies in the dataset. }
    \label{tab:dataset}
\end{table}

\subsection{Evaluation metrics and statistical tests}
\label{subsec:em}

\subsubsection{Evaluation metrics}
\label{subsubsec:metrics}
We evaluate the effectiveness of \method with packet-level metrics in RQ1 - RQ4. We also demonstrate the practical implications of \method by reporting incident-level metrics in RQ1. Moreover, the perplexity metric is presented in RQ3 to illustrate the quality of LM. 
~\\
\noindent\textbf{Packet-level effectiveness metrics} evaluate the predictive performance of \method on a single packet. Following the common practices in the literature, we adopt three commonly used classification metrics: precision, recall, and F1 score. \textit{Precision} measures the accuracy of the positive predictions. \textit{Recall} measures the proportion of actual positive cases correctly classified by the model. \textit{F1 score} is the harmonic average of precision and recall, evaluating the model from both perspectives.
~\\
\noindent\textbf{Incident-level effectiveness metrics} aim to evaluate \method from a higher and more practical perspective by focusing on the effectiveness with the anomaly incidents as the minimum unit. We define the following four metrics.  

\textit{Packet-Incident Coverage $\mathcal{C}_{PI}$} calculates the percentage of abnormal packets identified in a single incident. Let $\mathcal{N}_{totalP}$ denote the total number of packets of an anomaly incident and $\mathcal{N}_{correctP}$ denote the number of packets correctly identified by \method. We formally define $\mathcal{C}_{PI}$ as in Equation \ref{eq:cpi}.
    \begin{equation}
        \label{eq:cpi}
        \mathcal{C}_{PI}=\frac{\mathcal{N}_{correctP}}{\mathcal{N}_{totalP}}
    \end{equation}
    
\textit{Incident Coverage $\mathcal{C}_I$} measures how many anomaly incidents are identified. We assume an incident is identified if at least half of the abnormal packets are correctly classified. We denote the number of identified and total incidents as $\mathcal{N}_{correctI}$ and $\mathcal{N}_{totalI}$. We formally define $\mathcal{C}_I$ in Equation \ref{eq:ci}.
    \begin{equation}
        \label{eq:ci}
        \mathcal{C}_I=\frac{\mathcal{N}_{correctI}}{\mathcal{N}_{totalI}}
    \end{equation}

\textit{Detection Time Rate $\mathcal{DTR}_I$} assesses the time percentage needed to detect an anomaly incident. Let $t_{s}$ be the starting time of an incident and $\hat{t}_{s}$ be the time that \method identifies the first abnormal packet correctly. We denote the time duration for an anomaly incident as $\mathcal{T}_{totalT}$. Formally, $\mathcal{TR}_I$ is  defined in Equation \ref{eq:ti}.
    \begin{equation}
        \label{eq:ti}
        \mathcal{DTR}_I= \frac{\hat{t}_{s}-t_{s}}{\mathcal{T}_{totalT}}
    \end{equation}
    
\textit{Root Mean Square Error of Anomaly Length ($RMSE_L$)} evaluates how well \method can predict the anomaly length. Let the length of anomaly incident $i$ be $L_i$ and the predicted incident length be $\hat{L}_i$, we formally define $RMSE_L$ as: 
    \begin{equation}
RMSE_L = \sqrt{\frac{1}{n} \sum_{i=1}^n (L_i - \hat{L}_i)^2}
\end{equation}

\noindent\textbf{Perplexity of LM}. Perplexity is commonly used to evaluate probabilistic models, particularly LMs ~\cite{das2020can}. In the context of an LM, perplexity measures the average likelihood of its predictions for the next token. Higher perplexity indicates the model has more confidence in its prediction and vice versa. We calculate perplexity as in Equation \ref{eq:ppl}, where $n$ denotes the sequence length and $w_i$ represents the $i$th token in the sequence.
\begin{equation}
\label{eq:ppl}
PPL = \exp \left( -\frac{1}{n} \sum_{i=1}^{n} \log P(w_i | w_1, \ldots, w_{i-1}) \right)
\end{equation}

\subsubsection{Statistical testing}
\label{subsubsec:testing}
Using deep learning models introduces randomness into \method, which might threaten our empirical study's validity. Therefore, we repeat each experiment 30 times and perform the Mann-Whitney U test with a significance level of 0.01 as suggested in \cite{Arcuri2011}.  Furthermore, we evaluate the A12 effect size of the improvement~\cite{Arcuri2011}. \textit{Method A} has a higher chance of getting better values if the A12 value is greater than 0.5 and vice versa. We consider the effect size in the range  $[0.56, 0.64)$ as \emph{Small}, $[0.64, 0.71)$ as \emph{Medium}, and $[0.71, 1]$ as \emph{Large}.

\subsection{Experiment settings and execution}
\label{subsec:es}

\method involves several hyper-parameters that potentially introduce human biases. We performed a 10-fold cross validation  for the hyper-parameter selection with less bias~\cite{browne2000cross}. The code is built with the Pytorch framework~\cite{paszke2017automatic}. All the experiments were performed on one node from a national, experimental, heterogeneous computational cluster called eX3. This node is equipped with 2x Intel Xeon Platinum 8186, 1x NVIDIA V100 GPUs. 
\setlength{\intextsep}{10pt}
\setlength{\textfloatsep}{0pt}
\setlength{\abovecaptionskip}{0pt}
\setlength{\belowcaptionskip}{0pt}
\begin{table}[htb]
    \centering
    \begin{tabular}{>{\columncolor{gray!20}}lc>{\columncolor{gray!20}}lc}
    \toprule
    \rowcolor{white}
         Hyper-parameter & Value & Hyper-parameter & Value   \\
         \midrule
         optimizer& AdamW & lr & 0.001\\
          betas & (0.9, 0.999) & Embedd size & 64 \\
           $h^{vE}$\; size & 32 & $h^{dtmE}$ \;size & 16  \\
          $h^{vD}$\; size & 32 & $h^{dtmD}$\;size  & 16  \\
         CNN out channels & 6 & batch size & 12 \\
        \bottomrule
    \end{tabular}
    \caption{Hyperparameter values in \method. $lr$ denotes the learning rate of the optimizer, and $betas$ are also arguments of AdamW~\cite{loshchilov2017decoupled}. $h^{vE}$, $h^{dtmE}$, $h^{vD}$ and $h^{dtmD}$ represent the hidden vector of VAE encoder, VAE decoder, DTM encoder, and DTM decoder, respectively.}
    \label{tab:params}
\end{table}

\section{Experiment Results}
\label{sec:experi_resul}
\subsection{RQ1-\method effectiveness}
\label{subsec:res_rq1}
\setlength{\intextsep}{10pt}
\setlength{\textfloatsep}{0pt}
\setlength{\abovecaptionskip}{0pt}
\setlength{\belowcaptionskip}{0pt}
\begin{figure}
    \centering
    \includegraphics[width=0.9\linewidth]{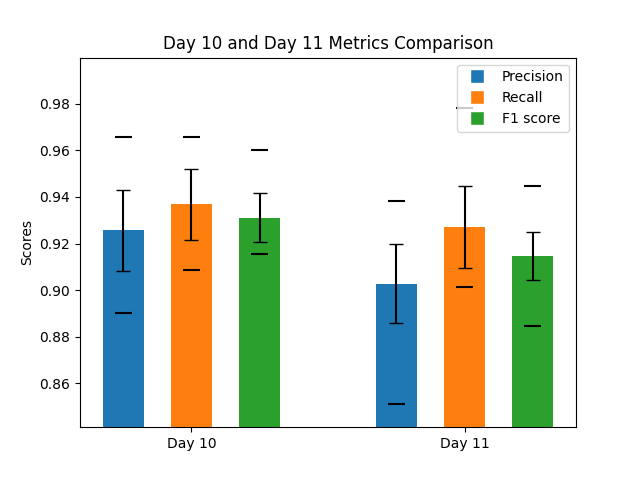}
    \caption{Results of the packet-level effectiveness metrics}
    \label{fig:barplot}
\end{figure}
Figure \ref{fig:barplot} delineates \method's performance on the packet level effectiveness metrics. The average F1 scores on both datasets are above 0.91, which demonstrates that \method is effective comprehensively. The precision results on both datasets are above 0.9, which represents more than 90\% of predicted anomalous packets are true anomalies. The recall results reach 0.927 on the Day 11 dataset and 0.937 on the Day 10 dataset, which indicates more than 92\% anomalous packets are successfully detected by \method.

We also observe that the performance of \method on the Day 11 dataset is generally inferior to that on the Day 10 dataset. One possible reason for this discrepancy is that the number of anomaly labels on the Day 11 dataset is smaller than that on the Day 10 dataset, indicating a higher training difficulty.
\setlength{\intextsep}{10pt}
\setlength{\textfloatsep}{0pt}
\setlength{\abovecaptionskip}{0pt}
\setlength{\belowcaptionskip}{0pt}
\begin{table}[htb]
    \centering
    \begin{tabular}{ccccc}
    \toprule
          Dataset & $\mathcal{C}_{PI}$ & $\mathcal{C}_{I}$ & $\mathcal{DTR}_I$ & $RMSE_L$ \\
          \midrule
        Day 10 & 87.36\% & 100\% & 3.81\% & 43.70 \\
        Day 11 & 81.09\% & 100\% & 6.22\% & 185.06 \\
        \bottomrule
    \end{tabular}
    \caption{Results of the incident-level effectiveness metrics}
    \label{tab:coarse}
\end{table}\\
Table \ref{tab:coarse} presents the experiment results for the incident-level effectiveness metrics. \textbf{$\mathcal{C}_{PI}$} results indicate that, on average, 87.36\% and 81.09\% abnormal packets in each anomaly incident are detected on Day 10 and Day 11, respectively. The incident coverages are both 100\%, showing that all incidents are successfully identified. \textbf{$\mathcal{DTR}_I$} on both datasets are relatively low ($<7\%$), implying that \method can detect an anomaly incident after the first 7\% packets. A low $\mathcal{DTR}_I$ indicates \method can detect anomalies near instantly after they take place, facilitating \textit{live} detection of various types of anomalies in TCMS, consequently curbing the potential damage to the whole system. \textbf{$RMSE_L$} signifies the standard deviation of the residuals, which represent the discrepancy between the predicted and the real incident duration. We can observe from the last column of Table \ref{tab:coarse} that the \textbf{$RMSE_L$} on both datasets (i.e., 43.70 and 185.06) is substantially lower than the average lengths of incidents (407.23 and 1127.52 as reported in Table \ref{tab:dataset}). This indicates that \method can predict the duration of an anomaly incident with a relatively small residual. The advantage of small residual can be harnessed for other applications, such as incident boundary determination, where a dedicated model can be established to predict the start and the end of an anomaly incident.
\noindent\cbox[gray!30]
{
\textbf{Concluding remarks on RQ1:} Regarding the packet-level effectiveness metrics (precision, recall and F1 score), \method achieves more than 90\% predictive performance on the Day 10 and Day 11 datasets. Regarding the incident-level effectiveness metrics, \method demonstrates high packet and incident coverages, low detection time rate, and relatively low RMSE of anomaly length prediction, implying \method's potential applications in incident detection, live detection, and incident boundary determination.}
\subsection{RQ2-DTM Effectiveness}
\label{subsec:res_rq2}
\method is a DT-based method, where DTM simulates TCMS and provides information about the system state to DTC. To evaluate the contribution of DTM, we compare \method and \method without DTM (denoted as \methodnodtm). 
\setlength{\intextsep}{10pt}
\setlength{\textfloatsep}{0pt}
\setlength{\abovecaptionskip}{0pt}
\setlength{\belowcaptionskip}{0pt}
\begin{table}[htb]
    \centering

    \begin{tabular}{c|lcccc}
\toprule
         Dataset &Metric &  \method &  NoDTM &  p-value &   A12 \\
\midrule
\multirow{3}{*}{Day 10} & Precision &    0.926 &  0.816 &    <0.01 & 1.000 \\
   & Recall &    0.937 &  0.802 &    <0.01 & 1.000 \\
    & F1 score &    0.931 &  0.809 &    <0.01 & 1.000 \\\hline
\multirow{3}{*}{Day 11} &Precision &    0.903 &  0.760 &    <0.01 & 1.000 \\
   & Recall &    0.927 &  0.820 &    <0.01 & 1.000 \\
       &F1 score &    0.915 &  0.789 &    <0.01 & 1.000 \\
\bottomrule
\end{tabular}
    \caption{Comparison of \method and \methodnodtm}
    \label{tab:rq2}
\end{table}\\
\noindent\cbox[gray!30]
{
\textbf{Concluding remarks on RQ2:} 
We observe a substantial decrease in the two datasets in terms of precision (12.7\%), recall (12.1\%), and F1 score (12.4\%) when we remove the DTM from \method. The DTM extracts inter-packet chronological features from the packets, which is indispensable for accurate anomaly detection.}
\subsection{RQ3-LM Effectiveness}
\begin{figure}
    \centering
    \includegraphics[width=0.87\linewidth]{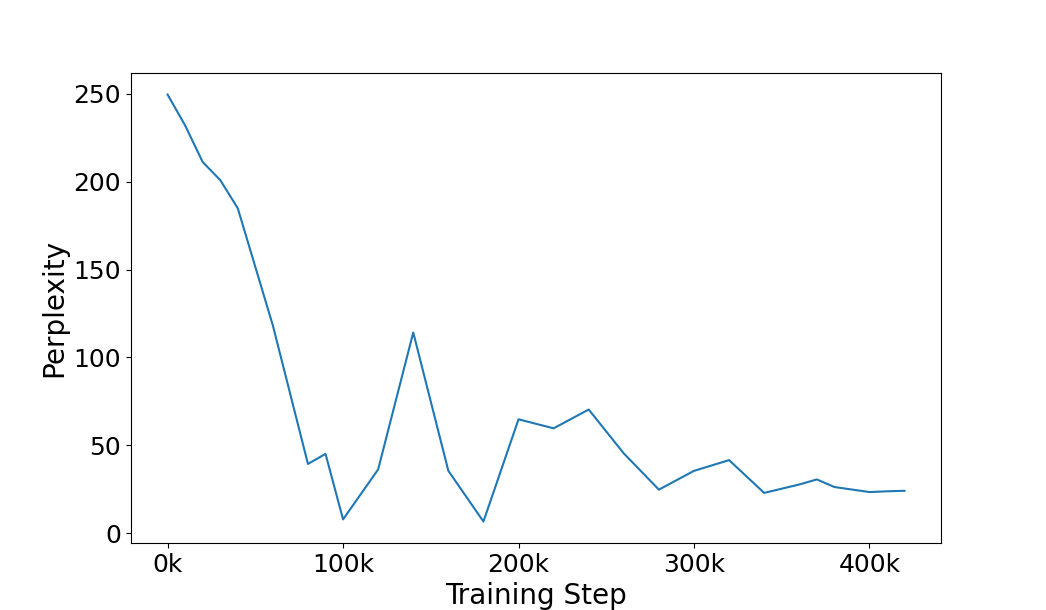}
    \caption{Results of perplexity during LM training}
    \label{fig:ppl}
\end{figure}

\method employs LM to extract inner-packet context features. To evaluate its effectiveness, we employ the perplexity metric (Section \ref{subsubsec:metrics}) to compare the performance of \method's LM with the baseline Word2vec model. Results are shown in Figure \ref{fig:ppl}. We can observe that the perplexity first decreases sharply and then gradually converges to around 30 at the end of training (after 400k batches).  Low perplexity indicates that the LM has high confidence in predicting the next token and extracting inner-packet context features. 

\label{subsec:res_rq3}

Table \ref{tab:rq3} shows the comparison results of \method and \methodwv. We can observe decreases in all three metrics on both datasets. The minimum decrease is $0.011$ ($0.926-0.915$) on the precision of the Day 10 dataset. The decreases are all significant except for the precision of the Day 10 dataset (p-value$=0.066>0.01$). We also observe a large effect size in recall and F1 score on both datasets ($A12>0.71$ as defined in Section \ref{subsubsec:testing}).

\begin{table}[hbt]
    \centering
    \begin{tabular}{c|lrrrr}
\toprule
          Dataset & Metric &  \method &  Word2Vec &  p-value &   A12 \\
\midrule
\multirow{3}{*}{Day 10} & Precision &    0.926 &     0.915 &    0.066 & 0.200 \\
   &Recall &    0.937 &     0.894 &    <0.01 & 0.733 \\
    & F1 score &    0.931 &     0.904 &    <0.01 & 0.867 \\\hline
\multirow{3}{*}{Day 11} & Precision  &    0.903 &     0.882 &    <0.01 & 0.600 \\
   & Recall &    0.927 &     0.882 &    <0.01 & 0.933 \\
   & F1 score &    0.915 &     0.882 &    <0.01 & 0.933 \\
\bottomrule
\end{tabular}
    \caption{Comparison of \method with \methodwv (denoted as Word2vec)}
    \label{tab:rq3}
\end{table}

\noindent\cbox[gray!30]
{
\textbf{Concluding remarks on RQ3:} 
LM presents low perplexity after training, suggesting strong feature extraction capability. \method outperforms \method-Word2vec significantly on both the Day 10 and Day 11 datasets (average F1 score improvement of 0.3).
}

\subsection{RQ4-Knowledge Distillation Effectiveness}
\label{subsec:res_rq4}
The training of \method is guided by the knowledge distilled from a teacher model, i.e., the VAE encoder. To demonstrate the effectiveness of KD, we compare \method and \method without KD  (denoted as \method-NoKD).
Table \ref{tab:rq4} depicts the comparison results. All metrics experienced a remarkable decrease with a minimum of 0.027 ($0.926-0.899$ on the precision of the Day 10 dataset). All the p-values in this table are smaller than 0.01, indicating the significance of the performance decrease brought by removing KD. We observe large effect sizes ($A12>0.71$) in all metrics except precision on Day 10 dataset, indicating \method-NoKD is highly probable to yield worse results than \method. Such results are consistent with our expectations as we posit KD can incorporate extra knowledge extracted from OOD data. Specifically, the pretrained VAE has a larger capacity in terms of model complexity compared to DTM, allowing it to fit a large $\mathcal{D}_{OOD}$ dataset without significant underfitting issues. The hidden vectors produced by the pretrained VAE epitomize its knowledge about feature extraction from a network packet. Therefore, using these hidden vectors as soft targets to guide the training of DTM entails a distillation process from the pretrained VAE to DTM. The extra knowledge distilled from the pretrained VAE supplements the training of DTM on the $\mathcal{D}_{ID}$ dataset, hence the predictive performance boost with KD.
\setlength{\intextsep}{10pt}
\setlength{\textfloatsep}{0pt}
\setlength{\abovecaptionskip}{0pt}
\setlength{\belowcaptionskip}{0pt}
 \begin{table}[htb]
    \centering
    \begin{tabular}{c|ccccc}
\toprule
          Dataset & Metric &  \method &  NoKD &  p-value &   A12 \\
\midrule
\multirow{3}{*}{Day 10} &Precision &    0.926 &       0.899 &    <0.01 & 0.333 \\
   & Recall &    0.937 &       0.874 &    <0.01 & 1.000 \\
       &F1 score &    0.931 &       0.886 &    <0.01 & 0.933 \\\hline
\multirow{3}{*}{Day 11} &Precision &    0.903 &       0.855 &    <0.01 & 0.933 \\
   &Recall &    0.927 &       0.824 &    <0.01 & 1.000 \\
       &F1 score &    0.915 &       0.839 &    <0.01 & 1.000 \\
\bottomrule
\end{tabular}
    \caption{Comparison of \method and \method-NoKD}
    \label{tab:rq4}
\end{table}

\noindent\cbox[gray!30]
{
\textbf{Concluding remarks on RQ4:} 
Removing KD from \method leads to a remarkable decrease ($>2.7\%$) in terms of average precision (3.75\%), recall (8.3\%), and F1 score (6.05\%). Most of the decreases are significant (p-value$<0.01$) with large effect sizes ($A12>0.71$). We conclude that KD is effective in improving the encoding capability of DTM.
}
\subsection{Threats to Validity}
\label{sec:threa}

\noindent\textbf{Construct Validity} concerns whether our chosen metrics accurately represent the anomaly detection quality. To be comprehensive, we include two sets of metrics: packet-level effectiveness metrics (i.e., precision, recall and F1 score) commonly used for evaluating anomaly detection methods and the incident-level effectiveness metrics (i.e., the packet-incident coverage, incident coverage, length RMSE and detection time rate) focusing on practical implications and assess \method from a domain perspective. We argue that these metrics together enable a more holistic assessment of \method.

\noindent\textbf{Internal Validity} refers to the credibility between cause and effect. One possible threat to internal validity is the selection of hyperparameters. The performance of DT may diminish in different settings. To reduce such threats, we choose these hyperparameters with cross-validation, which yields generic and optimal hyperparameters. The selection of hyperparameters does not introduce any human bias against the experiment datasets.

\noindent\textbf{Conclusion Validity} pertains to the validity of the conclusions drawn from the experiments. One common threat to conclusion validity is the randomness introduced in the model. \method harnesses neural networks for anomaly detection, inevitably introducing randomness. To mitigate this threat, we repeated each experiment 30 times. We performed statistical testing to study the significance of each improvement to ensure that the conclusions derived from our study are reliable. 

\noindent\textbf{External Validity} concerns the extent to which \method can generalize to other contexts. To reduce threats to external validity, we design our method to be generic, assuming no prior knowledge of the dataset distribution. Moreover, despite the high cost of collecting data from the real system, we obtained two separate datasets from different interfaces. Furthermore, we pretrain the VAE with the OOD dataset comprising anomaly-unrelated data samples, which is considered a universal task rather than bound to anomaly detection. The pretrained VAE pertains to knowledge of encoding a network packet into a hidden vector with high quality, benefitting various downstream tasks such as intrusion detection, network traffic analysis, and robustness analysis.  


\section{Practical Implications}
\label{sec:imp}
\textbf{Automating anomaly detection.} Our experiment results show that \method is effective for anomaly detection of \company's TCMS network, comprehensively measured with commonly used metrics for evaluating predictive performance (e.g., precision) at the packet level and our newly proposed incident-level metrics such as anomaly incident coverage. Especially, we observed that \method reaches a 100\% incident coverage for both datasets, indicating that all anomaly incidents can be successfully detected. The practical implication of this observation is that a high incident coverage can release domain experts from excessive manual work of pinpointing anomaly incidents from vast network packets. 

\noindent\textbf{Enabling live monitoring.} Moreover, \method exhibited a low detection time rate, critical for enabling live monitoring of the TCMS network by providing near-instant alerts and warnings. An automatic protocol can be established to react to these warnings and alerts and, in turn, prevents the anomaly's further influence on the TCMS. Furthermore, we observe a relatively low RMSE of the anomaly length predicted by \method. Low RMSE demonstrates the effectiveness of utilizing \method for anomaly boundary determination. An accurate boundary determination facilitates the TCMS to react accordingly and appropriately. For example, the TCMS can neglect a short-length anomaly, thanks to the network's re-transmission mechanism, whereas a long-length anomaly can lead to function failures or even system failures, requiring resources and reactions from the TCMS to mitigate the impact of the anomaly. 
 
\noindent\textbf{Leveraging DT for improving CPS dependability.} Our experiment results highlight that \method highly benefits from having DTM in the DT structure. First, by simulating the TCMS, DTM can provide valuable information about the operation state of the TCMS (e.g., unstable network states where frequent re-transmissions occur). Second, developing the DTM is cost-efficient since it is automatically developed to simulate the TCMS by predicting the next network packet (Section \ref{subsubsec:dtm}). Predicting the next packet requires encoding the packet into a high-quality hidden vector, which is a universal task that can benefit many downstream tasks, such as intrusion detection and network traffic analysis. In other words, we can reuse the same DTM to support different DTCs.


\noindent\textbf{Benefiting from language models}. Our results show that LM can provide contextualized features compared to Word2vec. Extracting features from text is extensively researched in NLP. 
Though we believe pretraining an LM with network packet data is sufficient and cost-efficient for the network anomaly detection tasks in the TCMS, the potential benefit of exploring Large Language Models (LLMs) such as Elmo~\cite{peters_deep_2018}, Bert~\cite{devlin_bert_2019} and GPT~\cite{radford_improving_nodate}, is intriguing. As Wei et al.~\cite{wei_emergent_2022} pointed out, LLMs show emergent capability, representing high predictive performance and sample efficiency on downstream tasks that small LMs do not possess. We are motivated to investigate the future application of LLMs with network packet data.

\noindent\textbf{Using KD to alleviate the need for abundant in-domain data.} \method employs KD to make use of OOD data. The nature of supervised machine learning requires abundant labelled data for its training, which is expensive to collect. In some cases, collecting merely ID dataset is non-trivial as well. For instance, collecting anomaly-related data in the TCMS network (denoted as $\mathcal{D}_{ID}$) requires manual examination and collection by domain experts with the help of an in-house signal analysis tool DCUTerm. Moreover, extra work is required for labelling the ID dataset. To alleviate the need for ID data, KD can distil knowledge from the OOD dataset to DTM. Incorporating the OOD dataset unleashes \method from the sole dependency of the ID dataset at a relatively low cost since the OOD dataset (i.e., any normal network packets) is easier to collect in the TCMS network.


\section{Related work}
\label{sec:relat}
We discuss the related works from two aspects: DT for CPS and network anomaly detection.

\noindent\textbf{Digital Twin for CPS.}
\label{subsec:rel_dt}
 DTs have been investigated to enhance the security and safety of CPSs. DT originates from the Apollo program for mission training and support~\cite{Rosen2015AboutTI}. Later, DT was generalized to accommodate various CPSs, such as water treatment plants and power grids~\cite{yue_understanding_2021,xu_digital_2021}. Eckhart and Ekelhart designed a rule-based DT dedicated to CPS intrusion detection~\cite{eckhart_digital_2019}, which constantly checks rules violations that a CPS must adhere to under normal conditions. Instead of relying on prior knowledge, Damjanovic-Behrendt~\cite{damjanovic-behrendt_digital_2018} proposed to build a machine learning-based DT to learn privacy-related features directly from automobile industry datasets, including both historical and real-time data.  

Yue et al. \cite{yue_understanding_2021} proposed a DT conceptual model and further decoupled the two sub-components of DT:  DTM and DTC. They define the DTM as a live replica of the CPS, while the DTC is the functionality of the DT. Xu et al. ~\cite{xu_digital_2021} realized this conceptual model by building the DTM as a timed automata machine and the DTC as a Generative Adversarial Network (GCN). Experimental results on public testbeds demonstrate the effectiveness of such decoupled DT structures. Latsou et al.~\cite{latsou_digital_2023} extended the concept of DT and proposed a multi-agent DT architecture for anomaly detection and bottleneck identification in complex manufacturing systems. In addition to DT construction for a specific CPS, Xu et al.~\cite{qinghua_fse_2023} and Lu et al.~\cite{chengjie2023evoclinical} introduced transfer learning for DT evolution to synchronize with the changes in CPSs and software systems, respectively. Despite the success of these methods, building a data-driven DT inevitably requires sufficient ID data. In contrast, \method adapts a VAE as the DTM structure and utilizes KD to distill knowledge from OOD data. Doing so significantly increases the data volume, improving the performance of DT.

\noindent\textbf{Network Anomaly detection.}
\label{subsec:ad} Many approaches exist for network anomaly detection~\cite{ahmed_survey_2016}, where this task is generally formulated as a classification task.
Early research on network anomaly detection originates in statistical or rule-based approaches. Machine learning methods were used to estimate the data distribution and employ statistical testing to detect anomalies~\cite{statistical}. A rule-based network anomaly detector was proposed in ~\cite{guha_network_2017} to detect predefined rule violations and consider them anomalies.

Statistical and rule-based approaches require extensive effort from domain experts. To mitigate such problems, neural networks have been widely explored for network anomaly detection due to their capability of automatic feature extraction \cite{hooshmand_network_2022}. Kwon et al.~\cite{cnn} evaluated three CNN architectures for network anomaly detection and concluded that shallow CNN performs best. Li et al.~\cite{li_intrusion_2017} divide network fields into categorical and numerical fields and extract features separately. The feature vectors are then converted into images with 8*8 grayscale pixels and fed into a CNN to extract features automatically. 

Few works have considered network packets as raw text and utilized NLP for feature extraction. Goodman et al.~\cite{goodman_packet2vec_2020} modified the Word2vec model and generated embeddings for each token in a network packet. The proposed method then takes these embeddings as input features to classify network traffics into benign or malicious.
Our method \method follows this research line and uses contextualized LM as an alternative to Word2vec. LM provides valuable context information that can help extract high-quality features from intricate packets. Our results show that the contextualized LM outperforms Word2vec and leads to high predictive performance.

\section{Conclusion and Future Work}
\label{sec:concl}
In this study, we introduced a digital twin (DT) based approach, referred to as \method, for tackling the anomaly detection task within Train Control and Management System (TCMS) networks. \method leverages a language model (LM) and an LSTM to extract inner-packet context and inter-packet chronological features. To capitalize on the information present in out-of-domain (OOD) data, we train a Variational Autoencoder with OOD data and use it to guide the training process of the digital twin model (DTM).  We evaluate \method with two datasets from \company. Experimental results show the effectiveness of \method regarding packet-level and incident-level metrics. We also investigate the individual contribution of each component. Experimental results demonstrate the effectiveness of DTM, LM and KD, which can be leveraged to address various tasks other than anomaly detection. We plan to explore more contextualized LM, such as ElMo, Bert, and GPT. We are particularly interested in the potential application of ChatGPT in this domain despite of obvious drawbacks mentioned in Section \ref{sec:threa}.

\section{Acknowlegements} 
\label{sec:ack}
The project is supported by the security project funded by the Norwegian Ministry of Education and Research, the Horizon 2020 project ADEPTNESS (871319) funded by the European Commission, and the Co-tester (\#314544) project funded by the Research Council of Norway (RCN). This work has benefited from the Experimental Infrastructure for Exploration of Exascale Computing (eX3), which is financially supported by RCN under contract 270053. 
\balance
\bibliographystyle{ACM-Reference-Format} 
\bibliography{references}


\begin{thebibliography}{40}


\ifx \showCODEN    \undefined \def \showCODEN     #1{\unskip}     \fi
\ifx \showDOI      \undefined \def \showDOI       #1{#1}\fi
\ifx \showISBNx    \undefined \def \showISBNx     #1{\unskip}     \fi
\ifx \showISBNxiii \undefined \def \showISBNxiii  #1{\unskip}     \fi
\ifx \showISSN     \undefined \def \showISSN      #1{\unskip}     \fi
\ifx \showLCCN     \undefined \def \showLCCN      #1{\unskip}     \fi
\ifx \shownote     \undefined \def \shownote      #1{#1}          \fi
\ifx \showarticletitle \undefined \def \showarticletitle #1{#1}   \fi
\ifx \showURL      \undefined \def \showURL       {\relax}        \fi
\providecommand\bibfield[2]{#2}
\providecommand\bibinfo[2]{#2}
\providecommand\natexlab[1]{#1}
\providecommand\showeprint[2][]{arXiv:#2}

\bibitem[Ahmed et~al\mbox{.}(2016)]%
        {ahmed_survey_2016}
\bibfield{author}{\bibinfo{person}{Mohiuddin Ahmed}, \bibinfo{person}{Abdun
  Naser~Mahmood}, {and} \bibinfo{person}{Jiankun Hu}.}
  \bibinfo{year}{2016}\natexlab{}.
\newblock \showarticletitle{A survey of network anomaly detection techniques}.
\newblock \bibinfo{journal}{\emph{Journal of Network and Computer
  Applications}}  \bibinfo{volume}{60} (\bibinfo{date}{Jan.}
  \bibinfo{year}{2016}), \bibinfo{pages}{19--31}.
\newblock
\showISSN{10848045}
\urldef\tempurl%
\url{https://doi.org/10.1016/j.jnca.2015.11.016}
\showDOI{\tempurl}


\bibitem[Anushiya and Lavanya(2021)]%
        {anushiya_comparative_2021}
\bibfield{author}{\bibinfo{person}{R Anushiya} {and} \bibinfo{person}{V~S
  Lavanya}.} \bibinfo{year}{2021}\natexlab{}.
\newblock \showarticletitle{A {COMPARATIVE} {STUDY} {ON} {INTRUSION}
  {DETECTION} {SYSTEMS} {FOR} {SECURED} {COMMUNICATION} {IN} {INTERNET} {OF}
  {THINGS}}.
\newblock \bibinfo{journal}{\emph{ICTACT JOURNAL ON COMMUNICATION TECHNOLOGY}}
  \bibinfo{volume}{12}, \bibinfo{number}{03} (\bibinfo{year}{2021}).
\newblock


\bibitem[Arcuri and Briand(2011)]%
        {Arcuri2011}
\bibfield{author}{\bibinfo{person}{Andrea Arcuri} {and} \bibinfo{person}{Lionel
  Briand}.} \bibinfo{year}{2011}\natexlab{}.
\newblock \showarticletitle{A practical guide for using statistical tests to
  assess randomized algorithms in software engineering}.
\newblock \bibinfo{journal}{\emph{Proceedings - International Conference on
  Software Engineering}} (\bibinfo{year}{2011}), \bibinfo{pages}{1--10}.
\newblock
\showISBNx{9781450304450}
\showISSN{02705257}
\urldef\tempurl%
\url{https://doi.org/10.1145/1985793.1985795}
\showDOI{\tempurl}


\bibitem[Browne(2000)]%
        {browne2000cross}
\bibfield{author}{\bibinfo{person}{Michael~W Browne}.}
  \bibinfo{year}{2000}\natexlab{}.
\newblock \showarticletitle{Cross-validation methods}.
\newblock \bibinfo{journal}{\emph{Journal of mathematical psychology}}
  \bibinfo{volume}{44}, \bibinfo{number}{1} (\bibinfo{year}{2000}),
  \bibinfo{pages}{108--132}.
\newblock


\bibitem[Clark et~al\mbox{.}(2018)]%
        {clark2018semi}
\bibfield{author}{\bibinfo{person}{Kevin Clark}, \bibinfo{person}{Minh-Thang
  Luong}, \bibinfo{person}{Christopher~D Manning}, {and}
  \bibinfo{person}{Quoc~V Le}.} \bibinfo{year}{2018}\natexlab{}.
\newblock \showarticletitle{Semi-supervised sequence modeling with cross-view
  training}.
\newblock \bibinfo{journal}{\emph{arXiv preprint arXiv:1809.08370}}
  (\bibinfo{year}{2018}).
\newblock


\bibitem[Damjanovic-Behrendt(2018)]%
        {damjanovic-behrendt_digital_2018}
\bibfield{author}{\bibinfo{person}{Violeta Damjanovic-Behrendt}.}
  \bibinfo{year}{2018}\natexlab{}.
\newblock \bibinfo{booktitle}{\emph{A {Digital} {Twin}-based {Privacy}
  {Enhancement} {Mechanism} for the {Automotive} {Industry}}}.
\newblock
\urldef\tempurl%
\url{https://doi.org/10.1109/IS.2018.8710526}
\showDOI{\tempurl}
\newblock
\shownote{Pages: 279}.


\bibitem[Das and Verma(2020)]%
        {das2020can}
\bibfield{author}{\bibinfo{person}{Avisha Das} {and} \bibinfo{person}{Rakesh~M
  Verma}.} \bibinfo{year}{2020}\natexlab{}.
\newblock \showarticletitle{Can Machines Tell Stories? A Comparative Study of
  Deep Neural Language Models and Metrics.}
\newblock \bibinfo{journal}{\emph{IEEE Access}}  \bibinfo{volume}{8}
  (\bibinfo{year}{2020}), \bibinfo{pages}{181258--181292}.
\newblock


\bibitem[Devlin et~al\mbox{.}(2019)]%
        {devlin_bert_2019}
\bibfield{author}{\bibinfo{person}{Jacob Devlin}, \bibinfo{person}{Ming-Wei
  Chang}, \bibinfo{person}{Kenton Lee}, {and} \bibinfo{person}{Kristina
  Toutanova}.} \bibinfo{year}{2019}\natexlab{}.
\newblock \bibinfo{title}{{BERT}: {Pre}-training of {Deep} {Bidirectional}
  {Transformers} for {Language} {Understanding}}.
\newblock
\newblock
\urldef\tempurl%
\url{http://arxiv.org/abs/1810.04805}
\showURL{%
\tempurl}
\newblock
\shownote{arXiv:1810.04805 [cs]}.


\bibitem[Dong et~al\mbox{.}(2021)]%
        {DONG2021100379}
\bibfield{author}{\bibinfo{person}{Shi Dong}, \bibinfo{person}{Ping Wang},
  {and} \bibinfo{person}{Khushnood Abbas}.} \bibinfo{year}{2021}\natexlab{}.
\newblock \showarticletitle{A survey on deep learning and its applications}.
\newblock \bibinfo{journal}{\emph{Computer Science Review}}
  \bibinfo{volume}{40} (\bibinfo{year}{2021}), \bibinfo{pages}{100379}.
\newblock
\showISSN{1574-0137}
\urldef\tempurl%
\url{https://doi.org/10.1016/j.cosrev.2021.100379}
\showDOI{\tempurl}


\bibitem[Eckhart and Ekelhart(2019)]%
        {eckhart_digital_2019}
\bibfield{author}{\bibinfo{person}{Matthias Eckhart} {and}
  \bibinfo{person}{Andreas Ekelhart}.} \bibinfo{year}{2019}\natexlab{}.
\newblock \showarticletitle{Digital {Twins} for {Cyber}-{Physical} {Systems}
  {Security}: {State} of the {Art} and {Outlook}}.
\newblock In \bibinfo{booktitle}{\emph{Security and {Quality} in
  {Cyber}-{Physical} {Systems} {Engineering}: {With} {Forewords} by {Robert}
  {M}. {Lee} and {Tom} {Gilb}}}, \bibfield{editor}{\bibinfo{person}{Stefan
  Biffl}, \bibinfo{person}{Matthias Eckhart}, \bibinfo{person}{Arndt Lüder},
  {and} \bibinfo{person}{Edgar Weippl}} (Eds.). \bibinfo{publisher}{Springer
  International Publishing}, \bibinfo{address}{Cham},
  \bibinfo{pages}{383--412}.
\newblock
\showISBNx{978-3-030-25312-7}
\urldef\tempurl%
\url{https://doi.org/10.1007/978-3-030-25312-7_14}
\showDOI{\tempurl}


\bibitem[Eskin(2000)]%
        {statistical}
\bibfield{author}{\bibinfo{person}{Eleazar Eskin}.}
  \bibinfo{year}{2000}\natexlab{}.
\newblock \showarticletitle{Anomaly Detection over Noisy Data Using Learned
  Probability Distributions}. In \bibinfo{booktitle}{\emph{Proceedings of the
  Seventeenth International Conference on Machine Learning}}
  \emph{(\bibinfo{series}{ICML '00})}. \bibinfo{publisher}{Morgan Kaufmann
  Publishers Inc.}, \bibinfo{address}{San Francisco, CA, USA},
  \bibinfo{pages}{255–262}.
\newblock
\showISBNx{1558607072}


\bibitem[Ghosh et~al\mbox{.}(2017)]%
        {guha_network_2017}
\bibfield{author}{\bibinfo{person}{Soumadip Ghosh}, \bibinfo{person}{Arindrajit
  Pal}, \bibinfo{person}{Amitava Nag}, \bibinfo{person}{Shayak Sadhu}, {and}
  \bibinfo{person}{Ramsekher Pati}.} \bibinfo{year}{2017}\natexlab{}.
\newblock \showarticletitle{Network anomaly detection using a fuzzy rule-based
  classifier}.
\newblock In \bibinfo{booktitle}{\emph{Computer, {Communication} and
  {Electrical} {Technology}} (\bibinfo{edition}{1} ed.)},
  \bibfield{editor}{\bibinfo{person}{Debatosh Guha}, \bibinfo{person}{Badal
  Chakraborty}, {and} \bibinfo{person}{Himadri Sekhar~Dutta}} (Eds.).
  \bibinfo{publisher}{CRC Press}, \bibinfo{pages}{61--65}.
\newblock
\showISBNx{978-1-315-40062-4}
\urldef\tempurl%
\url{https://doi.org/10.1201/9781315400624-12}
\showDOI{\tempurl}


\bibitem[Goodman et~al\mbox{.}(2020)]%
        {goodman_packet2vec_2020}
\bibfield{author}{\bibinfo{person}{Eric~L. Goodman}, \bibinfo{person}{Chase
  Zimmerman}, {and} \bibinfo{person}{Corey Hudson}.}
  \bibinfo{year}{2020}\natexlab{}.
\newblock \bibinfo{title}{{Packet2Vec}: {Utilizing} {Word2Vec} for {Feature}
  {Extraction} in {Packet} {Data}}.
\newblock
\newblock
\urldef\tempurl%
\url{http://arxiv.org/abs/2004.14477}
\showURL{%
\tempurl}
\newblock
\shownote{arXiv:2004.14477 [cs]}.


\bibitem[Guzman et~al\mbox{.}(2018)]%
        {guzman_data-driven_2018}
\bibfield{author}{\bibinfo{person}{Daniela~Narezo Guzman},
  \bibinfo{person}{Edin Hadzic}, \bibinfo{person}{Robert Schuil},
  \bibinfo{person}{Eric Baars}, {and} \bibinfo{person}{Jörn~Christoffer
  Groos}.} \bibinfo{year}{2018}\natexlab{}.
\newblock \showarticletitle{Data-driven condition now- and forecasting of
  railway switches for improvement in the quality of railway transportation}.
\newblock  (\bibinfo{year}{2018}).
\newblock


\bibitem[{H. Zhao} et~al\mbox{.}(2017)]%
        {h_zhao_fault_2017}
\bibfield{author}{\bibinfo{person}{{H. Zhao}}, \bibinfo{person}{{H. Chen}},
  \bibinfo{person}{{W. Dong}}, \bibinfo{person}{{X. Sun}}, {and}
  \bibinfo{person}{{Y. Ji}}.} \bibinfo{year}{2017}\natexlab{}.
\newblock \showarticletitle{Fault diagnosis of rail turnout system based on
  case-based reasoning with compound distance methods}. In
  \bibinfo{booktitle}{\emph{2017 29th {Chinese} {Control} {And} {Decision}
  {Conference} ({CCDC})}}. \bibinfo{pages}{4205--4210}.
\newblock
\showISBNx{1948-9447}
\urldef\tempurl%
\url{https://doi.org/10.1109/CCDC.2017.7979237}
\showDOI{\tempurl}
\newblock
\shownote{Journal Abbreviation: 2017 29th Chinese Control And Decision
  Conference (CCDC)}.


\bibitem[Han et~al\mbox{.}(2022a)]%
        {liping-tosem}
\bibfield{author}{\bibinfo{person}{Liping Han}, \bibinfo{person}{Shaukat Ali},
  \bibinfo{person}{Tao Yue}, \bibinfo{person}{Aitor Arrieta}, {and}
  \bibinfo{person}{Maite Arratibel}.} \bibinfo{year}{2022}\natexlab{a}.
\newblock \bibinfo{booktitle}{\emph{{Uncertainty-aware Robustness Assessment of
  Industrial Elevator Systems}}}.
\newblock \bibinfo{type}{{T}echnical {R}eport}.
\newblock


\bibitem[Han et~al\mbox{.}(2022b)]%
        {liping}
\bibfield{author}{\bibinfo{person}{Liping Han}, \bibinfo{person}{Tao Yue},
  \bibinfo{person}{Shaukat Ali}, \bibinfo{person}{Aitor Arrieta}, {and}
  \bibinfo{person}{Maite Arratibel}.} \bibinfo{year}{2022}\natexlab{b}.
\newblock \showarticletitle{Are Elevator Software Robust against Uncertainties?
  Results and Experiences from an Industrial Case Study}. In
  \bibinfo{booktitle}{\emph{Proceedings of the 30th ACM Joint European Software
  Engineering Conference and Symposium on the Foundations of Software
  Engineering}} (Singapore, Singapore) \emph{(\bibinfo{series}{ESEC/FSE
  2022})}. \bibinfo{publisher}{Association for Computing Machinery},
  \bibinfo{address}{New York, NY, USA}, \bibinfo{pages}{1331–1342}.
\newblock
\showISBNx{9781450394130}
\urldef\tempurl%
\url{https://doi.org/10.1145/3540250.3558955}
\showDOI{\tempurl}


\bibitem[Hooshmand and Hosahalli(2022)]%
        {hooshmand_network_2022}
\bibfield{author}{\bibinfo{person}{Mohammad~Kazim Hooshmand} {and}
  \bibinfo{person}{Doreswamy Hosahalli}.} \bibinfo{year}{2022}\natexlab{}.
\newblock \showarticletitle{Network anomaly detection using deep learning
  techniques}.
\newblock \bibinfo{journal}{\emph{CAAI Transactions on Intelligence
  Technology}} \bibinfo{volume}{7}, \bibinfo{number}{2} (\bibinfo{date}{June}
  \bibinfo{year}{2022}), \bibinfo{pages}{228--243}.
\newblock
\showISSN{2468-2322, 2468-2322}
\urldef\tempurl%
\url{https://doi.org/10.1049/cit2.12078}
\showDOI{\tempurl}


\bibitem[Islam et~al\mbox{.}(2022)]%
        {islam_novel_2022}
\bibfield{author}{\bibinfo{person}{Umar Islam}, \bibinfo{person}{Rami~Qays
  Malik}, \bibinfo{person}{Amnah~S. Al-Johani}, \bibinfo{person}{Muhammad.~Riaz
  Khan}, \bibinfo{person}{Yousef~Ibrahim Daradkeh}, \bibinfo{person}{Ijaz
  Ahmad}, \bibinfo{person}{Khalid~A. Alissa}, \bibinfo{person}{Zulkiflee
  Abdul-Samad}, {and} \bibinfo{person}{Elsayed~M. Tag-Eldin}.}
  \bibinfo{year}{2022}\natexlab{}.
\newblock \showarticletitle{A {Novel} {Anomaly} {Detection} {System} on the
  {Internet} of {Railways} {Using} {Extended} {Neural} {Networks}}.
\newblock \bibinfo{journal}{\emph{Electronics}} \bibinfo{volume}{11},
  \bibinfo{number}{18} (\bibinfo{date}{Sept.} \bibinfo{year}{2022}),
  \bibinfo{pages}{2813}.
\newblock
\showISSN{2079-9292}
\urldef\tempurl%
\url{https://doi.org/10.3390/electronics11182813}
\showDOI{\tempurl}


\bibitem[Jones et~al\mbox{.}(2020)]%
        {jones2020characterising}
\bibfield{author}{\bibinfo{person}{David Jones}, \bibinfo{person}{Chris
  Snider}, \bibinfo{person}{Aydin Nassehi}, \bibinfo{person}{Jason Yon}, {and}
  \bibinfo{person}{Ben Hicks}.} \bibinfo{year}{2020}\natexlab{}.
\newblock \showarticletitle{Characterising the Digital Twin: A systematic
  literature review}.
\newblock \bibinfo{journal}{\emph{CIRP Journal of Manufacturing Science and
  Technology}}  \bibinfo{volume}{29} (\bibinfo{year}{2020}),
  \bibinfo{pages}{36--52}.
\newblock


\bibitem[Kwon et~al\mbox{.}(2018)]%
        {cnn}
\bibfield{author}{\bibinfo{person}{Donghwoon Kwon}, \bibinfo{person}{Kathiravan
  Natarajan}, \bibinfo{person}{Sang~C. Suh}, \bibinfo{person}{Hyunjoo Kim},
  {and} \bibinfo{person}{Jinoh Kim}.} \bibinfo{year}{2018}\natexlab{}.
\newblock \showarticletitle{An Empirical Study on Network Anomaly Detection
  Using Convolutional Neural Networks}. In \bibinfo{booktitle}{\emph{2018 IEEE
  38th International Conference on Distributed Computing Systems (ICDCS)}}.
  \bibinfo{pages}{1595--1598}.
\newblock
\urldef\tempurl%
\url{https://doi.org/10.1109/ICDCS.2018.00178}
\showDOI{\tempurl}


\bibitem[Lamping and Warnicke(2004)]%
        {lamping2004wireshark}
\bibfield{author}{\bibinfo{person}{Ulf Lamping} {and} \bibinfo{person}{Ed
  Warnicke}.} \bibinfo{year}{2004}\natexlab{}.
\newblock \showarticletitle{Wireshark user's guide}.
\newblock \bibinfo{journal}{\emph{Interface}} \bibinfo{volume}{4},
  \bibinfo{number}{6} (\bibinfo{year}{2004}), \bibinfo{pages}{1}.
\newblock


\bibitem[Latsou et~al\mbox{.}(2023)]%
        {latsou_digital_2023}
\bibfield{author}{\bibinfo{person}{Christina Latsou}, \bibinfo{person}{Maryam
  Farsi}, {and} \bibinfo{person}{John~Ahmet Erkoyuncu}.}
  \bibinfo{year}{2023}\natexlab{}.
\newblock \showarticletitle{Digital twin-enabled automated anomaly detection
  and bottleneck identification in complex manufacturing systems using a
  multi-agent approach}.
\newblock \bibinfo{journal}{\emph{Journal of Manufacturing Systems}}
  \bibinfo{volume}{67} (\bibinfo{date}{April} \bibinfo{year}{2023}),
  \bibinfo{pages}{242--264}.
\newblock
\showISSN{0278-6125}
\urldef\tempurl%
\url{https://doi.org/10.1016/j.jmsy.2023.02.008}
\showDOI{\tempurl}


\bibitem[Li et~al\mbox{.}(2017)]%
        {li_intrusion_2017}
\bibfield{author}{\bibinfo{person}{Zhipeng Li}, \bibinfo{person}{Zheng Qin},
  \bibinfo{person}{Kai Huang}, \bibinfo{person}{Xiao Yang}, {and}
  \bibinfo{person}{Shuxiong Ye}.} \bibinfo{year}{2017}\natexlab{}.
\newblock \showarticletitle{Intrusion {Detection} {Using} {Convolutional}
  {Neural} {Networks} for {Representation} {Learning}}. In
  \bibinfo{booktitle}{\emph{Neural {Information} {Processing}}},
  \bibfield{editor}{\bibinfo{person}{Derong Liu}, \bibinfo{person}{Shengli
  Xie}, \bibinfo{person}{Yuanqing Li}, \bibinfo{person}{Dongbin Zhao}, {and}
  \bibinfo{person}{El-Sayed~M. El-Alfy}} (Eds.). \bibinfo{publisher}{Springer
  International Publishing}, \bibinfo{address}{Cham},
  \bibinfo{pages}{858--866}.
\newblock
\showISBNx{978-3-319-70139-4}


\bibitem[Loshchilov and Hutter(2017)]%
        {loshchilov2017decoupled}
\bibfield{author}{\bibinfo{person}{Ilya Loshchilov} {and}
  \bibinfo{person}{Frank Hutter}.} \bibinfo{year}{2017}\natexlab{}.
\newblock \showarticletitle{Decoupled weight decay regularization}.
\newblock \bibinfo{journal}{\emph{arXiv preprint arXiv:1711.05101}}
  (\bibinfo{year}{2017}).
\newblock


\bibitem[Lu et~al\mbox{.}(2023)]%
        {chengjie2023evoclinical}
\bibfield{author}{\bibinfo{person}{Chengjie Lu}, \bibinfo{person}{Qinghua Xu},
  \bibinfo{person}{Tao Yue}, \bibinfo{person}{Shaukat Ali},
  \bibinfo{person}{Thomas Schwitalla}, {and} \bibinfo{person}{Jan~F. Nygård}.}
  \bibinfo{year}{2023}\natexlab{}.
\newblock \showarticletitle{EvoCLINICAL: Evolving Cyber-cyber Digital Twin with
  Active Transfer Learning for Automated Cancer Registry System}. In
  \bibinfo{booktitle}{\emph{Proceedings of the 31th ACM Joint European Software
  Engineering Conference and Symposium on the Foundations of Software
  Engineering}} (San Francisco, CA, USA) \emph{(\bibinfo{series}{ESEC/FSE
  2023})}. \bibinfo{publisher}{Association for Computing Machinery},
  \bibinfo{address}{New York, NY, USA}, \bibinfo{numpages}{11}~pages.
\newblock
\showISBNx{979-8-4007-0327-0/23/12}
\urldef\tempurl%
\url{https://doi.org/10.1145/3611643.3613897}
\showDOI{\tempurl}


\bibitem[Mimura and Tanaka(2018)]%
        {mimura2018reading}
\bibfield{author}{\bibinfo{person}{Mamoru Mimura} {and} \bibinfo{person}{Hidema
  Tanaka}.} \bibinfo{year}{2018}\natexlab{}.
\newblock \showarticletitle{Reading network packets as a natural language for
  intrusion detection}. In \bibinfo{booktitle}{\emph{Information Security and
  Cryptology--ICISC 2017: 20th International Conference, Seoul, South Korea,
  November 29-December 1, 2017, Revised Selected Papers 20}}. Springer,
  \bibinfo{pages}{339--350}.
\newblock


\bibitem[Paszke et~al\mbox{.}(2017)]%
        {paszke2017automatic}
\bibfield{author}{\bibinfo{person}{Adam Paszke}, \bibinfo{person}{Sam Gross},
  \bibinfo{person}{Soumith Chintala}, \bibinfo{person}{Gregory Chanan},
  \bibinfo{person}{Edward Yang}, \bibinfo{person}{Zachary DeVito},
  \bibinfo{person}{Zeming Lin}, \bibinfo{person}{Alban Desmaison},
  \bibinfo{person}{Luca Antiga}, {and} \bibinfo{person}{Adam Lerer}.}
  \bibinfo{year}{2017}\natexlab{}.
\newblock \showarticletitle{Automatic differentiation in pytorch}.
\newblock  (\bibinfo{year}{2017}).
\newblock


\bibitem[Peters et~al\mbox{.}(2018)]%
        {peters_deep_2018}
\bibfield{author}{\bibinfo{person}{Matthew~E. Peters}, \bibinfo{person}{Mark
  Neumann}, \bibinfo{person}{Mohit Iyyer}, \bibinfo{person}{Matt Gardner},
  \bibinfo{person}{Christopher Clark}, \bibinfo{person}{Kenton Lee}, {and}
  \bibinfo{person}{Luke Zettlemoyer}.} \bibinfo{year}{2018}\natexlab{}.
\newblock \bibinfo{title}{Deep contextualized word representations}.
\newblock
\newblock
\urldef\tempurl%
\url{http://arxiv.org/abs/1802.05365}
\showURL{%
\tempurl}
\newblock
\shownote{arXiv:1802.05365 [cs]}.


\bibitem[Pinheiro~Cinelli et~al\mbox{.}(2021)]%
        {pinheiro_cinelli_variational_2021}
\bibfield{author}{\bibinfo{person}{Lucas Pinheiro~Cinelli},
  \bibinfo{person}{Matheus Araújo~Marins}, \bibinfo{person}{Eduardo~Antúnio
  Barros~da Silva}, {and} \bibinfo{person}{Sérgio Lima~Netto}.}
  \bibinfo{year}{2021}\natexlab{}.
\newblock \showarticletitle{Variational {Autoencoder}}.
\newblock In \bibinfo{booktitle}{\emph{Variational {Methods} for {Machine}
  {Learning} with {Applications} to {Deep} {Networks}}},
  \bibfield{editor}{\bibinfo{person}{Lucas~Pinheiro Cinelli},
  \bibinfo{person}{Matheus~Araújo Marins}, \bibinfo{person}{Eduardo~Antônio
  Barros~da Silva}, {and} \bibinfo{person}{Sérgio~Lima Netto}} (Eds.).
  \bibinfo{publisher}{Springer International Publishing},
  \bibinfo{address}{Cham}, \bibinfo{pages}{111--149}.
\newblock
\showISBNx{978-3-030-70679-1}
\urldef\tempurl%
\url{https://doi.org/10.1007/978-3-030-70679-1_5}
\showDOI{\tempurl}


\bibitem[Radford et~al\mbox{.}({[n.\,d.]})]%
        {radford_improving_nodate}
\bibfield{author}{\bibinfo{person}{Alec Radford}, \bibinfo{person}{Karthik
  Narasimhan}, \bibinfo{person}{Tim Salimans}, {and} \bibinfo{person}{Ilya
  Sutskever}.} \bibinfo{year}{[n.\,d.]}\natexlab{}.
\newblock \showarticletitle{Improving {Language} {Understanding} by
  {Generative} {Pre}-{Training}}.
\newblock  (\bibinfo{year}{[n.\,d.]}).
\newblock


\bibitem[Rosen et~al\mbox{.}(2015)]%
        {Rosen2015AboutTI}
\bibfield{author}{\bibinfo{person}{Roland Rosen}, \bibinfo{person}{Georg von
  Wichert}, \bibinfo{person}{George Lo}, {and} \bibinfo{person}{Kurt~Dirk
  Bettenhausen}.} \bibinfo{year}{2015}\natexlab{}.
\newblock \showarticletitle{About The Importance of Autonomy and Digital Twins
  for the Future of Manufacturing}.
\newblock \bibinfo{journal}{\emph{IFAC-PapersOnLine}}  \bibinfo{volume}{48}
  (\bibinfo{year}{2015}), \bibinfo{pages}{567--572}.
\newblock


\bibitem[Wei et~al\mbox{.}(2022)]%
        {wei_emergent_2022}
\bibfield{author}{\bibinfo{person}{Jason Wei}, \bibinfo{person}{Yi Tay},
  \bibinfo{person}{Rishi Bommasani}, \bibinfo{person}{Colin Raffel},
  \bibinfo{person}{Barret Zoph}, \bibinfo{person}{Sebastian Borgeaud},
  \bibinfo{person}{Dani Yogatama}, \bibinfo{person}{Maarten Bosma},
  \bibinfo{person}{Denny Zhou}, \bibinfo{person}{Donald Metzler},
  \bibinfo{person}{Ed~H. Chi}, \bibinfo{person}{Tatsunori Hashimoto},
  \bibinfo{person}{Oriol Vinyals}, \bibinfo{person}{Percy Liang},
  \bibinfo{person}{Jeff Dean}, {and} \bibinfo{person}{William Fedus}.}
  \bibinfo{year}{2022}\natexlab{}.
\newblock \bibinfo{title}{Emergent {Abilities} of {Large} {Language} {Models}}.
\newblock
\newblock
\urldef\tempurl%
\url{http://arxiv.org/abs/2206.07682}
\showURL{%
\tempurl}
\newblock
\shownote{arXiv:2206.07682 [cs]}.


\bibitem[Xu et~al\mbox{.}(2021)]%
        {xu_digital_2021}
\bibfield{author}{\bibinfo{person}{Qinghua Xu}, \bibinfo{person}{Shaukat Ali},
  {and} \bibinfo{person}{Tao Yue}.} \bibinfo{year}{2021}\natexlab{}.
\newblock \showarticletitle{Digital twin-based anomaly detection in
  cyber-physical systems}. \bibinfo{publisher}{IEEE},
  \bibinfo{pages}{205--216}.
\newblock
\showISBNx{1-72816-836-8}


\bibitem[Xu et~al\mbox{.}(2023a)]%
        {xu_digital_2023}
\bibfield{author}{\bibinfo{person}{Qinghua Xu}, \bibinfo{person}{Shaukat Ali},
  {and} \bibinfo{person}{Tao Yue}.} \bibinfo{year}{2023}\natexlab{a}.
\newblock \showarticletitle{Digital {Twin}-based {Anomaly} {Detection} with
  {Curriculum} {Learning} in {Cyber}-physical {Systems}}.
\newblock \bibinfo{journal}{\emph{ACM Transactions on Software Engineering and
  Methodology}} (\bibinfo{date}{Feb.} \bibinfo{year}{2023}),
  \bibinfo{pages}{3582571}.
\newblock
\showISSN{1049-331X, 1557-7392}
\urldef\tempurl%
\url{https://doi.org/10.1145/3582571}
\showDOI{\tempurl}


\bibitem[Xu et~al\mbox{.}(2022)]%
        {xu_uncertainty-aware_2022}
\bibfield{author}{\bibinfo{person}{Qinghua Xu}, \bibinfo{person}{Shaukat Ali},
  \bibinfo{person}{Tao Yue}, {and} \bibinfo{person}{Maite Arratibel}.}
  \bibinfo{year}{2022}\natexlab{}.
\newblock \showarticletitle{Uncertainty-aware transfer learning to evolve
  digital twins for industrial elevators}. In
  \bibinfo{booktitle}{\emph{Proceedings of the 30th {ACM} {Joint} {European}
  {Software} {Engineering} {Conference} and {Symposium} on the {Foundations} of
  {Software} {Engineering}}}. \bibinfo{publisher}{ACM},
  \bibinfo{address}{Singapore Singapore}, \bibinfo{pages}{1257--1268}.
\newblock
\showISBNx{978-1-4503-9413-0}
\urldef\tempurl%
\url{https://doi.org/10.1145/3540250.3558957}
\showDOI{\tempurl}


\bibitem[Xu et~al\mbox{.}(2023b)]%
        {qinghua_fse_2023}
\bibfield{author}{\bibinfo{person}{Qinghua Xu}, \bibinfo{person}{Shaukat Ali},
  \bibinfo{person}{Tao Yue}, \bibinfo{person}{Nedim Zaimovic}, {and}
  \bibinfo{person}{Singh Inderjeet}.} \bibinfo{year}{2023}\natexlab{b}.
\newblock \showarticletitle{Uncertainty-Aware Transfer Learning to Evolve
  Digital Twins for Industrial Elevators}. In
  \bibinfo{booktitle}{\emph{Proceedings of the 31th ACM Joint European Software
  Engineering Conference and Symposium on the Foundations of Software
  Engineering}} (San Francisco, CA, USA) \emph{(\bibinfo{series}{ESEC/FSE
  2023})}. \bibinfo{publisher}{Association for Computing Machinery},
  \bibinfo{address}{New York, NY, USA}, \bibinfo{numpages}{11}~pages.
\newblock
\showISBNx{979-8-4007-0327-0/23/12}
\urldef\tempurl%
\url{https://doi.org/10.1145/3611643.3613879}
\showDOI{\tempurl}


\bibitem[Yu et~al\mbox{.}(2019)]%
        {yu2019review}
\bibfield{author}{\bibinfo{person}{Yong Yu}, \bibinfo{person}{Xiaosheng Si},
  \bibinfo{person}{Changhua Hu}, {and} \bibinfo{person}{Jianxun Zhang}.}
  \bibinfo{year}{2019}\natexlab{}.
\newblock \showarticletitle{A review of recurrent neural networks: LSTM cells
  and network architectures}.
\newblock \bibinfo{journal}{\emph{Neural computation}} \bibinfo{volume}{31},
  \bibinfo{number}{7} (\bibinfo{year}{2019}), \bibinfo{pages}{1235--1270}.
\newblock


\bibitem[Yue et~al\mbox{.}(2021)]%
        {yue_understanding_2021}
\bibfield{author}{\bibinfo{person}{Tao Yue}, \bibinfo{person}{Paolo Arcaini},
  {and} \bibinfo{person}{Shaukat Ali}.} \bibinfo{year}{2021}\natexlab{}.
\newblock \showarticletitle{Understanding {Digital} {Twins} for
  {Cyber}-{Physical} {Systems}: {A} {Conceptual} {Model}}. In
  \bibinfo{booktitle}{\emph{Leveraging {Applications} of {Formal} {Methods},
  {Verification} and {Validation}: {Tools} and {Trends}}},
  \bibfield{editor}{\bibinfo{person}{Tiziana Margaria} {and}
  \bibinfo{person}{Bernhard Steffen}} (Eds.). \bibinfo{publisher}{Springer
  International Publishing}, \bibinfo{address}{Cham}, \bibinfo{pages}{54--71}.
\newblock
\showISBNx{978-3-030-83723-5}


\bibitem[Zafar et~al\mbox{.}(2021)]%
        {zafar_model-based_2021}
\bibfield{author}{\bibinfo{person}{Muhammad~Nouman Zafar},
  \bibinfo{person}{Wasif Afzal}, \bibinfo{person}{Eduard Enoiu},
  \bibinfo{person}{Athanasios Stratis}, \bibinfo{person}{Aitor Arrieta}, {and}
  \bibinfo{person}{Goiuria Sagardui}.} \bibinfo{year}{2021}\natexlab{}.
\newblock \showarticletitle{Model-{Based} {Testing} in {Practice}: {An}
  {Industrial} {Case} {Study} using {GraphWalker}}. In
  \bibinfo{booktitle}{\emph{14th {Innovations} in {Software} {Engineering}
  {Conference} (formerly known as {India} {Software} {Engineering}
  {Conference})}}. \bibinfo{publisher}{ACM}, \bibinfo{address}{Bhubaneswar,
  Odisha India}, \bibinfo{pages}{1--11}.
\newblock
\showISBNx{978-1-4503-9046-0}
\urldef\tempurl%
\url{https://doi.org/10.1145/3452383.3452388}
\showDOI{\tempurl}


\end{thebibliography}
\end{document}